\documentclass{article}

\usepackage{arxiv}

\usepackage[utf8]{inputenc} % allow utf-8 input
\usepackage[T1]{fontenc}    % use 8-bit T1 fonts
\usepackage{hyperref}       % hyperlinks
\usepackage{url}            % simple URL typesetting
\usepackage{booktabs}       % professional-quality tables
\usepackage{amsfonts}       % blackboard math symbols
\usepackage{nicefrac}       % compact symbols for 1/2, etc.
\usepackage{microtype}      % microtypography
\usepackage{lipsum}		% Can be removed after putting your text content
\usepackage{graphicx}
\usepackage{hyperref}
\usepackage{natbib}
\usepackage{doi}
\usepackage{rotating}
\usepackage{longtable}
\usepackage{makecell}
\usepackage[]{mdframed}
\usepackage{lineno}
\usepackage{marvosym}
\usepackage{multirow}
\usepackage{pdflscape} 
\usepackage{array}
\usepackage[capitalise,nameinlink]{cleveref}
\usepackage{subcaption}

\defcitealias{pandas_development_team_pandas_2022}{Pandas development team, 2022}
\defcitealias{python_software_foundation_python_2022}{Python software foundation, 2022}
\defcitealias{r_core_team_introduction_2024}{R Core team, 2024}

\title{Machine learning applications in archaeological practices: a review}

\author{ \href{https://orcid.org/0000-0003-0319-1562}{\includegraphics[scale=0.06]{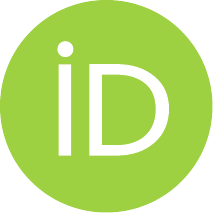}\hspace{1mm}Mathias.~Bellat}\thanks{Shared first authorship.} \\
	Collaborative Research Center 1070 “ResourceCultures”\\
 	University of Tübingen, Tübingen, 72074, Germany\\
    Department of Geosciences, Working Group Soil Science and Geomorphology\\
	University of Tübingen, Tübingen, 72074, Germany\\
	\texttt{mathias.bellat@uni-tuebingen.de} \\
	\And
	\href{https://orcid.org/0000-0002-2814-6306}{\includegraphics[scale=0.06]{orcid.pdf}\hspace{1mm}Jordy D.~Orellana Figueroa*} \\
	Department of Geosciences, Working Group Early Prehistory and Quaternary Ecology\\
	University of Tübingen,Tübingen, 72074, Germany\\
    High Performance and Cloud Computing Group\\
    University of Tübingen,	Tübingen, 72074, Germany\\
	\texttt{ext-contact@jorellanaf.com} \\
 	\And
	\href{https://orcid.org/0000-0002-2017-0539}{\includegraphics[scale=0.06]{orcid.pdf}\hspace{1mm}Jonathan S.~Reeves} \\
	Department of Geosciences, Working Group Early Prehistory and Quaternary Ecology\\
	University of Tübingen,	Tübingen, 72074, Germany\\
    Technological Primates Research Group\\
    Max Planck Institute for Evolutionary Anthropology, Leipzig, 04103, Germany\\
  	\And
	\href{https://orcid.org/0000-0002-4620-6624}{\includegraphics[scale=0.06]{orcid.pdf}\hspace{1mm}Ruhollah ~Taghizadeh-Mehrjardi} \\
    Faculty of Agriculture and Natural Resources \\
    Ardakan University, Ardakan, Iran\\
  	\And
	\href{https://orcid.org/0000-0002-5302-4925}{\includegraphics[scale=0.06]{orcid.pdf}\hspace{1mm}Claudio~Tennie}\thanks{Shared senior authorship.} \\
	Department of Geosciences, Working Group Early Prehistory and Quaternary Ecology\\
	University of Tübingen, Tübingen, 72070, Germany\\
      	\And
	\href{https://orcid.org/0000-0002-4875-2602}{\includegraphics[scale=0.06]{orcid.pdf}\hspace{1mm}Thomas~Scholten} \Cross. \\
	Collaborative Research Center 1070 “ResourceCultures”\\
  	University of Tübingen,	Tübingen, 72074, Germany\\
    Department of Geosciences, Working Group Soil Science and Geomorphology\\
    University of Tübingen,	Tübingen, 72074, Germany\\
    DFG Cluster of Excellence “Machine Learning: New Perspectives for Science”\\
	University of Tübingen,	Tübingen, 72076, Germany\\
}

\renewcommand{\headeright}{Accepted preprint}
\renewcommand{\shorttitle}{Machine learning applications in archaeological practices: a review}

\hypersetup{
pdftitle={Machine learning applications in archaeological practices: a review},
pdfsubject={cs.LG},
pdfauthor={Mathias.~Bellat, Jordy.~D. Orellana~Figueroa},
pdfkeywords={archaeological sciences, computer sciences, state of art, artificial intelligence, machine learning},
}

\begin{document}
\maketitle
\clearpage
\begin{abstract}
Artificial intelligence and machine learning applications in archaeology have increased significantly in recent years, and these now span all subfields, geographical regions, and time periods from prehistory to modern times. The prevalence and success of these applications have remained largely unexamined, as recent reviews on the use of machine learning in archaeology have only focused only on specific subfields of archaeology. Our review examined an exhaustive corpus of 135 articles published between 1997 and 2022. We observed a significant increase in the number of relevant publications from 2019 onwards, mainly concentrated in a few journals and mainly published in open access format. Automatic structure detection and artefact classification were the most represented tasks in the articles reviewed, followed by taphonomy, archaeological predictive modelling, and architectural classification or reconstruction. From the corpus of articles analysed, clustering and unsupervised methods were underrepresented compared to supervised models. Artificial neural networks and ensemble learning account for two thirds of the total number of models used. However, if machine learning models are gaining in popularity they remain subject to misunderstanding. We observed, in some cases, poorly defined requirements and caveats of the machine learning methods used. Furthermore, the goals and the needs of machine learning applications for archaeological purposes are in some cases unclear or poorly expressed. To address this, we propose here a workflow guide for archaeologists to develop coherent and consistent methodologies adapted to their research questions, project scale and available data. As in many areas of modern life, machine learning is rapidly becoming an important tool in archaeological research and practice, particularly useful for the analyses of large and highly multivariate data, although not without limitations. This review highlights the importance of well-defined and well-reported structured methodologies and collaborative practices to maximise the potential of applications of machine learning methods in archaeological research.
\end{abstract}

% keywords can be removed

\section{Introduction}
\label{1}
A study by Binford and Binford [\citeyear{binford_preliminary_1966}] on the characterisation and classification of Mousterian assemblages was one of the first steps into multivariate statistical analysis in archaeological research, and opened a window for further, more developed applications. Other major developments in statistical and computing analyses soon followed \citep[p.66]{djindjian_short_2015}, dealing with classification and clustering problems \citep{hodson_cluster_1970}, predictive archaeology \citep{judge_quantifying_1988}, or various quantitative techniques to fully bring together data structures, quantitative analyses and theoretical interpretations \citep{carr_for_1989,voorrips_mathematics_1990}. The first traces or machine learning applications in archaeology already date back to the 1970s \citep{kowalski_classification_1972, thomas_empirical_1973}, and their extensive use has grown with advancements in computing technology \citep{damico_machine_2022, cardarelli_legacy_2025}.

\par Unlike traditional statistical models, like linear models such as ordinary least squares or general linear models that find solutions with formal equations, machine learning methods seek to use the data itself to create a model able to accurately evaluate new, unseen data. Machine learning is defined by Alpaydin as  “\textit{programming computers to optimize a performance criterion using example data or past experience}” \citep[p.3]{alpaydin_introduction_2014}. Machine learning models require a training phase, where the computer learns from the data to improve its model and thus its predictions. Once trained, machine learning models often yield faster, more accurate results than traditional methods, all while lowering the costs [\textit{e.g.} \citealt{kochkov_neural_2024}]. This benefit can make machine learning an appealing solution to certain scientific questions \citep{verhagen_case_2007,hansen_prioritizing_2020,calder_use_2022, brandsen_labelling_2022, orellana_figueroa_virtual_nodate}, Orellana Figueroa \textit{et al., in press}. However, this comes at the cost of both the models being difficult to interpret \citep{carvalho_machine_2019}, and the training process generally requiring large amounts of (high quality) data, time, and compute power; in a way, front-loading the time and costs to the training process to produce fast and accurate predictions once trained \citep{sevilla_compute_2022}.

\par In archaeology, machine learning includes a variety of applications such as the classification of zooarchaeological remains \citep{boon_digital_2009, anichini_automatic_2021, cole_evaluating_2022, anglisano_supervised_2022}, spatial pattern and mobility analyses \citep{stott_searching_2019, stular_migration_2022}, cultural heritage reconstruction and preservation \citep{toler-franklin_multi-feature_2010, grilli_classification_2019, castiello_computational_2022, parsons_hard-hearted_2023}, the study of settlement dynamics \citep{miera_large-scale_2022}, taphonomic classification \citep{byeon_automated_2019} and on-site analysis of the origin and function of sediments and the spatial distribution of artefacts \citep{orengo_brave_2019, ginau_what_2020, reese_deep_2021, agapiou_detection_2021}.

\par The frequency, extent, and success of machine learning applications across archaeology remain largely unknown and their implications and ethical considerations are under debate \citep{davis_defining_2020, bickler_machine_2021, tenzer_debating_2024, gattiglia_managing_2025, neri_ethical_2025}. Previous reviews of machine learning applications in archaeology mainly focused on remote sensing \citep{jamil_review_2022, argyrou_review_2022, bellat_humans_2025}, text analysis \citep{gonzalez-perez_discourse_2023},  classification of ceramics \citep{ling_findings_2024}, wear-use analysis \citep{eleftheriadou_machine_2025} and other artefacts \citep{naso_state_2025}, site-conservation problematics \citep{casillo_artificial_2025}, or only on a small set of study cases \citep{mantovan_computerization_2020}. On the other hand, while Cacciari and Pocobelli [\citeyear{damico_machine_2022}] have assessed machine learning applications across a wide range of fields of archaeological research, they mainly provide qualitative observations rather than a quantitative analysis as presented here, as is also the case for the overview published by Calder et al. [\citeyear{calder_use_2022}]. Palacios [\citeyear{palacios_aplicacion_2023}] provided an innovative introduction on the status of machine learning in archaeology, although mainly focused on Bayesian approaches, and with a limited selection of reviewed papers mainly on archaeological predictive models. A recent comparison of machine learning methods to PCA and LDA for species identity from faunal material find that methods like RF and LDA outperform PCA \citep{cole_evaluating_2022}.

\par The inherent importance of the fieldwork component that generates the data for machine learning approaches, as well as the variety of methods and practices existing in archaeology \citep{kelly_archaeology_2017, renfrew_archaeology_2020} all provide some difficulties in delimiting the scope of the field. Statistical analyses of artefact morphology and their typological classification, the process of site prospection using satellite imagery, the quantitative analysis of animal bones at a human-created site, and the analysis of the soil and geological composition of a site are all part of archaeological research.

\par Moreover, machine learning refers to a large set of different statistical methods and algorithms, from those based on Bayesian statistics, to evolutionary algorithms that seek to emulate the process of natural selection \citep{hastie_elements_2009, kubat_introduction_2017}. Recent developments in artificial neural networks have had an impact far beyond the limits of academic research \citep{pangti_machine_2021, bachute_autonomous_2021, kawamleh_algorithmic_2024}, not to mention the emergence of so-called generative artificial intelligence, which is trained to create text, images, audio, or even video based on simple text prompts \citep{zhang_stackgan_2016, vaswani_attention_2017, liu_generative_2023}. Such advances in machine learning methods and their successful application to a specific question in one area often led to experimentation in other, sometimes completely unrelated, areas. The breadth of methods and practices used in archaeology thus provides many opportunities for new advances in machine learning to be applied.

\par Therefore, if we wish to obtain a comprehensive view of how machine learning methods have impacted the field of archaeology, we must look broadly across the various machine learning architectures, as well as the many subfields and chronological focuses of archaeological research, from the survey phase to post-excavation analysis and interpretation. Furthermore, to determine the frequency of machine learning applications and whether the recent high-profile advances in artificial intelligence have increased interest in its use in archaeological research, we must undertake a quantitative and chronological analysis of the published literature.

\par With this review article, we aim to provide a comprehensive analysis of previous applications of machine learning methods in the field of archaeology, our primary objective is to provide an overview of current trends and methodological pitfalls in the field. As secondary aims, we propose potential solutions to these issues and outline possible future directions for research. We analysed a corpus of academic publications across a – as comprehensive as possible – set of categories of both machine learning method families and different archaeological subfields, presenting a chronological overview of machine learning applications in a wide spectrum of the archaeological research, as well as a discussion of our analysis of the reviewed articles, the methods applied, the results obtained, and the provided conclusions.

\section{Methods}
\label{methods}
A rapid systematic review protocol was conducted \citep{petticrew_systematic_2006, grant_typology_2009, jesson_doing_2012, peters_guidance_2015, page_prisma_2021} to obtain a representative dataset for a broad overview of machine learning applications in archaeological research.  Although there is no consensus on the specific methodology of a rapid review, we followed the Preferred Reporting Items for Systematic Reviews and Meta-Analyses 2020 guideline (PRISMA, \citealt{page_prisma_2021}) and made our methodology as transparent as possible, as suggested by Haby et al. [\citeyear{haby_what_2016}].

\subsection{Search strategy}
The search strategy comprised two different and independent protocols: one performed for the automatic screening protocol  (considering a fully automated and scripted workflow), consisting of multiple searches, and one performed for the manual screening protocol (mixing automated and human selection steps), consisting of a single search [\textbf{\cref{tab1}, \cref{fig1}}]. The searches for the automatic protocol were performed by M.B and consisted of twelve queries each containing a combination of keywords [\textbf{\hyperref[box:1]{Box 1}}]. These keywords, while not covering the entire scope of possible archaeological study cases using machine learning (such as those not using the exact term of "machine learning"), already grasp a wide overview of machine learning applications in the field of archaeology. It is however important to note that this might have biased the search in favour of articles that use the keywords of “machine learning”. On the other hand, an earlier search using the manual protocol involved multiple queries using the 10 machine learning method families we used for our data analysis [see \textbf{\hyperref[annexe]{supplementary file}}], but these returned a large number of duplicate results, most of which were already captured by simply using "machine learning". The searches were performed in five online portals. All searches were conducted in English, except for the German National Library, where the search was conducted in German with only six of the queries [\textbf{\hyperref[box:1]{Box 1}}]. These searches covered all records published before 1 January 2023. In total, 1460 records were retrieved from the automated search protocol, of which 558 were unique.

\par The manual protocol search extracted results from \href{https://scholar.google.com/}{\textit{Google Scholar}}  without any constraints on years of publication, but was done on 6 January 2023 to ensure results only from 2022 and earlier. It was performed by J.D.O.F. and consisted of a single query of keyword combinations performed using a script written in \texttt{Python 3.9.16} \citepalias{python_software_foundation_python_2022} with the \texttt{habanero, pandas}, and \texttt{scholarly} libraries [\citealt{chamberlain_habanero_2022, cholewiak_scholarly_2022} \citetalias{pandas_development_team_pandas_2022}], along with \texttt{geckodriver} \citep{mozilla_geckodriver_2023} that obtained the top 300 results as ranked by Google Scholar, automatically parsing the metadata of these results \citep{orellana_figueroa_google_2020}. This search query and its keywords included the term “machine learning” and various forms of the term “archaeology” [\textbf{\hyperref[box:2]{Box 2}}] and retrieved a total of 285 unique records.

\begin{mdframed}\textit{Query 1= “archaeology machine learning” (“archäologie maschinelles lernen”)\\
Query 2 = “archeology machine learning”\\
Query 3 = “archaeological machine learning” (“archäologisch maschinelles lernen”)\\
Query 4 = “archeological machine learning”\\
Query 5 = “archaeology deep learning” (“archäologie deep learning”)\\
Query 6 = “archeology deep learning”\\
Query 7 = “archaeological deep learning” (“archäologisch deep lernen”)\\
Query 8 = “archeological deep learning”\\
Query 9 = “archaeology artificial intelligence” (“archäologie künstliche intelligenz”)\\
Query 10 = “archeology artificial intelligence”\\
Query 11 = “archaeological artificial intelligence” (“archäologisch künstliche intelligenz”)\\
Query 12 = “archeological artificial intelligence”}
\end{mdframed}
\begin{center}
    Box 1: Search queries used for the automatic protocol search. The German version of the queries used for the German National Library (DNB) portals are specified in parentheses where applicable (due to some queries being only different in the use of a spelling variant in English for "archaeology", which is not relevant in German).
    \label{box:1}
\end{center}
\vspace{0.5cm}

\begin{mdframed}\textit{ Topic = "machine learning" | "archaeological" | "archeological" | "archaeology" | "archeology" | "archaeo" | "archeo"}
\end{mdframed}
\begin{center}
    Box 2: Search query used for the manual protocol search.
    \label{box:2}
\end{center}

\begin{table}[ht]
	\centering
	\begin{tabular}{cc}
		\toprule
		\textbf{Bibliographical database} & \textbf{Keywords matches} \\
		\midrule
        \multicolumn{2}{c}{Automatic screening} \tabularnewline
        \\
		Web of Science & 969  \\
		PubMed & 413 \\
        Tübingen Universitätbibliothek & 51       \\
        German Archaeological Institute & 24       \\
        German National Library & 3      \\
        Total unique & 558 \\
        \midrule
        \multicolumn{2}{c}{Manual screening} \tabularnewline
        \\
		Google Scholar & 300 \\
        Total unique & 285 \\
        \midrule
        Total & 1760      \\
        \textbf{Sum of unique totals} & \textbf{730} \\
		\bottomrule
	\end{tabular}
 \vspace{0.25cm}
        \caption{Summary of results of both automatic and manual protocol searches on the six online portals. Note that the “Sum of unique totals” refers only to the sum of the number of non-duplicate results from each search; however, there were many publications present in both searches (see below), which would further reduce the sum total of unique items.}
	\label{tab1}
\end{table}

\subsection{Screening}

\begin{figure}[h!]
	\centering
	\includegraphics[width=1\linewidth, trim = {1.5cm, 0cm, 0cm, 1cm}, clip]{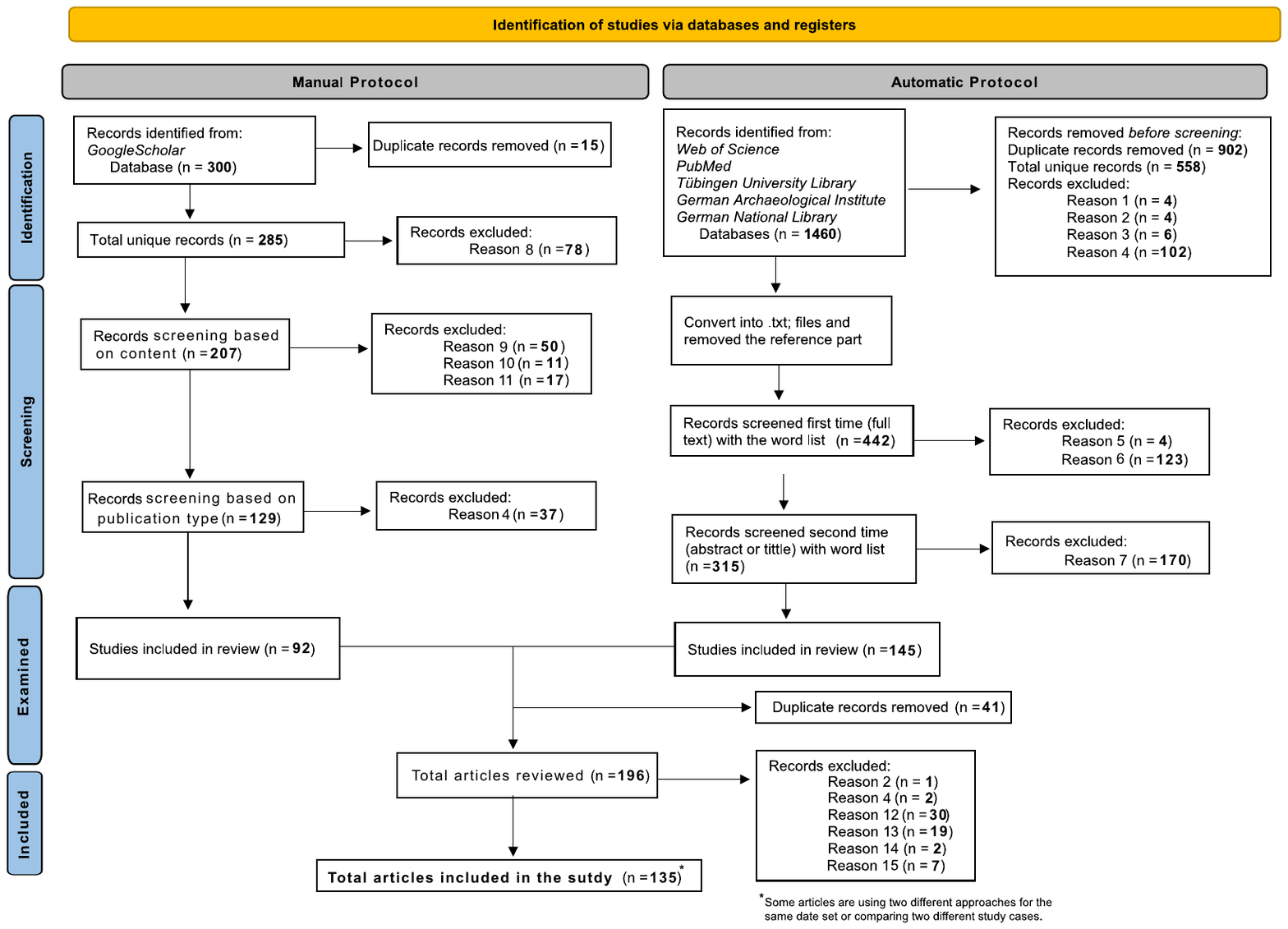}
	\caption{Review process from source selection to analysis. Inspired by the PRISMA 2020 flow diagram \citep{page_prisma_2021}. Reason 1 = Ineligible with automation tool; Reason 2 = Non-English record; Reason 3 = Full text not accessible; Reason 4 = Non-journal-based publications; Reason 5 = Absence of abstract; Reason 6 = Archaeology and machine learning keywords from the list not present in the text; Reason 7 = Archaeology and machine learning keywords from the list are not present in the abstract or in the title; Reason 8 = Preliminary exclusion (\textit{i.e.} no access to publication, publications or contribution by current authors, entire books, non-academic reports, preprints, potentially predatory journal; Reason 9 = Excluded based on the title; Reason 10 = Excluded based on abstract; Reason 11 = Excluded based on the full text first reading; Reason 12 = Full text does not involve archaeological research; Reason = 13 Full text does not involve machine learning methods as defined in our protocol; Reason 14 = Conflicts of interest (publication by the authors of this review or in which the authors contributed); Reason 15 = Theory or review paper. Figure created using Microsoft Word and Inkscape.}
	\label{fig1}
\end{figure}

Following the automatic and manual protocol searches, two independent screening procedures were performed on the retrieved records [\textbf{\cref{fig1}}]. For the automatic protocol, after excluding publications not written in English and records not published in scientific journals, the screening was performed with a script written in \texttt{R 4.4.1} \citepalias{r_core_team_introduction_2024} with the \texttt{corpus} and \texttt{tm} packages \citep{huang_corpus_2021,feinerer_tm_2023} divided in two filters. The first screening step retrieved all documents whose full text contained one of the archaeology or machine learning keywords we defined for this filtering step [\textbf{\hyperref[annexe]{supplementary file}}]. The second screening step was performed in the same manner with the same keywords, but using only the abstract and title of the results of the first screening step that were found to contain the keywords in the full text. A total of 135 records were included from the automatic search and screening protocol.

\par The manual screening protocol consisted of additional steps. First, a preliminary exclusion was performed, which checked whether the publications were inaccessible, whether they were preprints that were not yet in print, and whether the record was for an entire book, amongst other criteria [\textbf{\cref{fig1},{\hyperref[annexe]{supplementary file}}}]. Further screening was then performed based on the exclusion and inclusion criteria [\textbf{\hyperref[annexe]{supplementary file}}], firstly by examining the publication title alone. If after this, an article could not be securely excluded or included, a second screening step followed, now examining the abstract as well. If this was yet still insufficient for a secure inclusion or exclusion, the full-text was examined. A total of 93 records were included from the manual protocol search and screening protocol.

\par Both screening strategies used the same inclusion and exclusion criteria [\textbf{\cref{fig1},{\hyperref[annexe]{supplementary file}}}], though the automatic screening script, by necessity, used only a simplified version that only involved the use of keywords, while the manual protocol search was done through the human-based analysis of text.

\par However, to further verify the level of agreement between the two different screening strategies, the same automatic screening was performed on the 285 collected records from the manual protocol search. The results of this screening are reported below [cf. \ref{3.1}].

\subsection{Data extraction}

\par
\begin{figure}[h!]
	\centering
	\includegraphics[width=0.8\linewidth, trim = {0cm, 0.5cm, 0cm, 0cm}, clip]{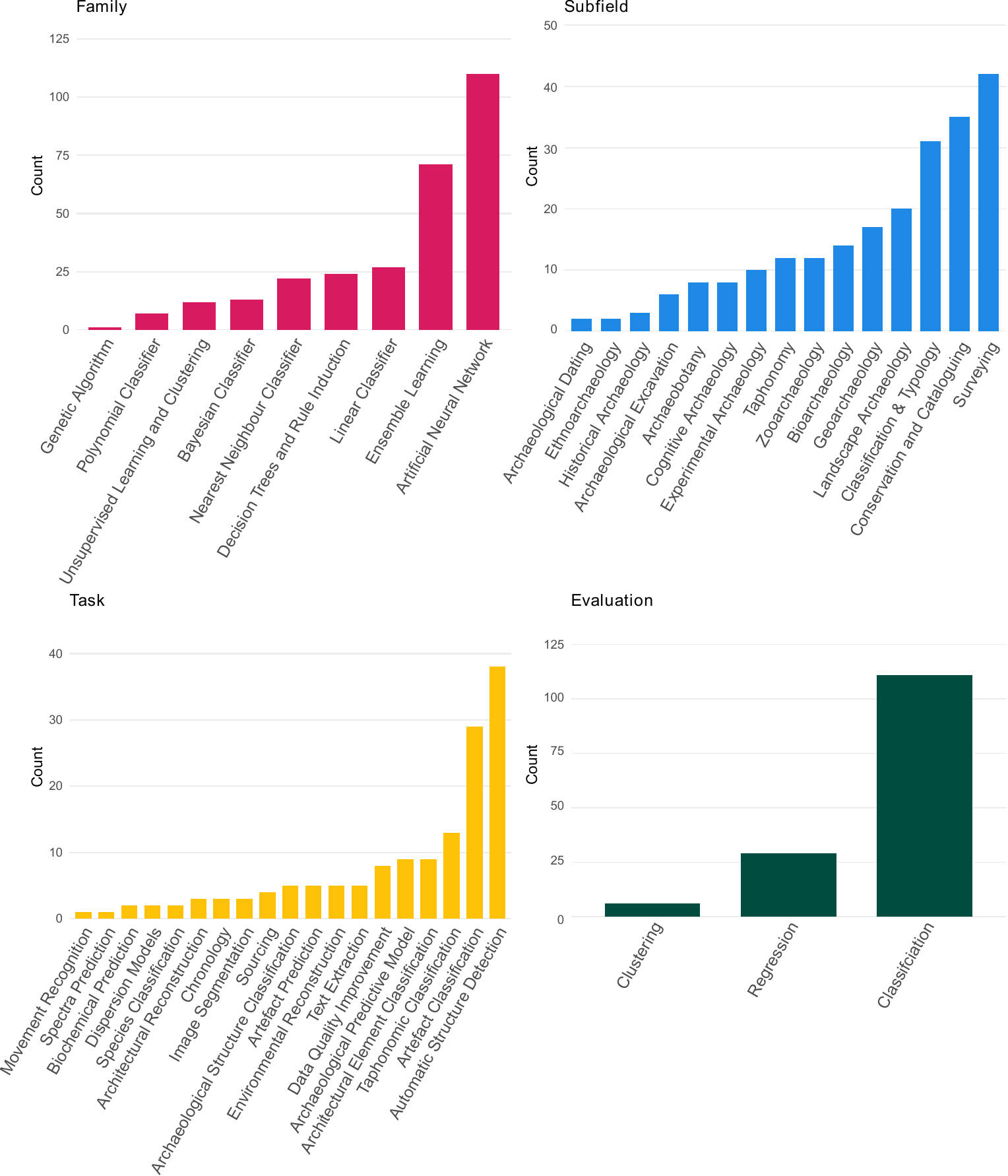}
\caption{The fourth field of information recorded in the review presents significant characteristics to explain variation in machine learning applications in archaeology and their related classes/categories. One study case might have been attributed to several subfields or architecture categories. Figure generated with \texttt{R 4.2.2} (code available in supplementary material 3) and additional editing with Inkscape.}
	\label{fig2}
\end{figure}

 We systematically divided each of the included articles into one or more study cases if different applications (\textit{e.g.} using different data or seeking different goals) were attested in the publication. Therefore, we obtained a total number of study cases (n = 147) superior to the sum of all included articles (n = 135). From all these study cases, we systematically recorded a total of nine features composed of different categories [\textbf{\cref{tab2}}]. These features were based on the authors' \textit{ad hoc} evaluation of possible features of interest in the evaluation of the study cases. No inter-rater reliability was calculated for the selection of these features. However, only four of these were needed to gain a clear overview of the practice of machine learning in archaeology: the architecture of the machine learning model used as well as its evaluation process (see below), the archaeological subfields where these methods were applied, and the type of task it was applied to [\textbf{\cref{fig2}}]. Though important, the other groups of features did not allow us to identify common patterns in the application of machine learning in archaeology. Thus, we included them only sporadically in our discussion. Furthermore, we extracted metadata related to the publication for our quantitative analysis. These metadata included the year of publication, the list of authors, the country of affiliation of the first author, the name of the journal, and whether the article was published through an open access modality.

\par Based on the model architecture, we divided machine learning methods into nine broad categories based on the family of algorithms they belonged to [\citealt{eleftheriadou_machine_2025}, \textbf{\cref{fig2}, {\hyperref[annexe]{supplementary file}}}]. We granularly recorded the different types of models used, which were then classified into the relevant architecture category, but we also recorded the frequency of their use. Each model was classified into only one architecture category, even in cases where multiple categories may have been warranted. For example, random forest is an ensemble of decision trees, fitting into both “ensemble learning” and “decision trees”, but was classified into “ensemble learning” only, as that was the more relevant aspect of the algorithm.\\
Furthermore, according to the goals of the model application, we grouped the evaluation into three broad categories: classification, regression, and clustering  [\textbf{\cref{fig2}}; \citealt[pp. 5-13]{alpaydin_introduction_2014}].

\begin{table}[ht]
	\centering
	\begin{tabular}{cc}
		\toprule
		\textbf{Features} & \textbf{Number of categories} \\
		\midrule
		Model & 70  \\
		Best model & 17 \\
        Family & 9  \\
        Subfield & 15  \\
        Input data & 11 \\
        Evaluation & 3 \\
	    Task & 19  \\
        Result & 5 \\
        Pre-training & 4 \\
		\bottomrule
	\end{tabular}
 \vspace{0.25cm}
        \caption{The nine features collected systematically from the review.}
	\label{tab2}
\end{table}

Due to the wide range of subjects, we divided the field of archaeology into fifteen subfields [\textbf{\cref{fig2}}] based on Kelly and Thomas [\citeyear{kelly_archaeology_2017}]. The assignment of a study case to a specific subfield was based on the authors’ own evaluation [\textbf{\hyperref[annexe]{supplementary file}}]. No inter-rater reliability calculation was performed for the assignment. A single publication could be classified into several subfields based on their research questions or objectives, and therefore, the sum total entries in the subfield feature does not equal the total number of study cases included in the review.\\
Finally, we classified every article’s application into nineteen \textit{a posteriori} categories based on the broader task (\textit{e.g.} automatic structure detection) for which the machine learning methods were applied [\textbf{\cref{fig2}}]. These task categories were \textit{ad hoc} and based on the authors’ evaluation of the corpus. Initially, we sought to define the tasks for which machine learning was applied for each publication from a very granular level, then aggregating the more granular task categories into broader ones a total of five times until we arrived at the nineteen categories reported here [\textbf{\hyperref[annexe]{supplementary file}}]. No inter-rater reliability calculation was performed for the creation of these tasks categories. These task categories were one of the most important characteristics to analyse, as these classes helped us understand not only the goals of the authors, but also the possible models that were more well-suited for the desired outcomes, based especially on previous applications.

\section{Results}
\subsection{Screening results}
\label{3.1}
Since the two screening protocols were performed independently, we will report them as such, even if the records obtained after both screenings were merged into a single set of publications [\textbf{\cref{fig1}}]. Out of the 558 unique records obtained from the automatic protocol, four could not be screened with our script (0.72\%, reason 1), another four were excluded, as they were not written in English (0.72\%, reason 2), and an additional six were also excluded as the full text was inaccessible (1.08\%, reason 3). Furthermore, 102 records were excluded, as they were not academic periodical journal articles (18.28\%, reason 4). From the remaining 442 records, an additional four were excluded, as they did not have any abstract available (0.72\%, reason 5). Furthermore, a screening based on the presence or absence of keywords [\textbf{\hyperref[annexe]{supplementary file}}] in the full-text was performed, and from this we excluded 123 records that did not include any of our defined keywords (27.83\%, reason 6). After the previous screening step, we performed a second filter on the 315 records which were left, based on the same set of keywords as previously but performed on the title and abstract only. A total of 170 (30.05\%, reason 7) additional records were then excluded, obtaining a total of 145 records for the final list of included publications from our automatic screening. With the 285 unique records from the manual query, we performed a preliminary exclusion and removed 78 records (27.37\%, reason 8). With the remaining 207 articles, we performed our title-based screening and excluded 50 records (17.54\%, reason 9), as well as our abstract-based screening, excluding an additional 11 records (3.86\%, reason 10). The final content-based screening, based on the full-text of the publications, led to the exclusion of 17 records (5.96\%, reason 11). After these screening steps we were left with 129 records. We then removed any articles that were not published in academic periodical journals, leading to the exclusion of 37 records (12.98\%, reason 4). We thus obtained the final list of included publications from our manual screening, containing a total of 92 items.\\
To verify the level of agreement between the two protocols, we performed the automatic screening protocol on the 285 unique records found through our manual protocol search. Of the 92 records included through the manual screening, 39 records were also included through automatic screening of the manual protocol search results, while 54 were included through manual screening but not through the automatic screening protocol. In addition, a further 25 records that were not included through manual screening were flagged positive through the automatic screening protocol, though these 25 were not part of the dataset of reviewed articles.

\par In total we obtained 196 unique items when both results from manual and automatic screening were merged. 42 articles were included through both screening processes, while 51 articles were unique to the manual protocol, and 103 were unique to the automatic one. Additional records were excluded \textit{a posteriori} during the reviewing of the articles for a wide variety of reasons [\textbf{\cref{fig1}}, reasons 12, 13, 14 and 15], though this was mostly relevant for those articles that were screened automatically, since the automatic screening script was not exact. In total, 61 records (31.12\% of 196) were excluded. The articles that were removed to avoid reviewing articles that the authors contributed to [\textbf{\cref{fig1}}, reason 14] were McPherron et al. [\citeyear{mcpherron_machine_2022}] and Orellana Figueroa et al. [\citeyear{orellana_figueroa_proof_2021}]. The final list of articles reviewed contained a total of 135 records.\\
Finally, to perform a simple diachronic analysis of the number of publications from recent years, we performed the same automatic screening and search protocol, but searching only for articles published between 1 January 2023 and 31 September 2024. The results are described below.

\subsection{Metadata analysis}
\begin{figure}[ht]
	\centering
	\includegraphics[width=0.6\linewidth, trim = {1.75cm, 5.5cm, 1.75cm, 5.5cm}, clip]{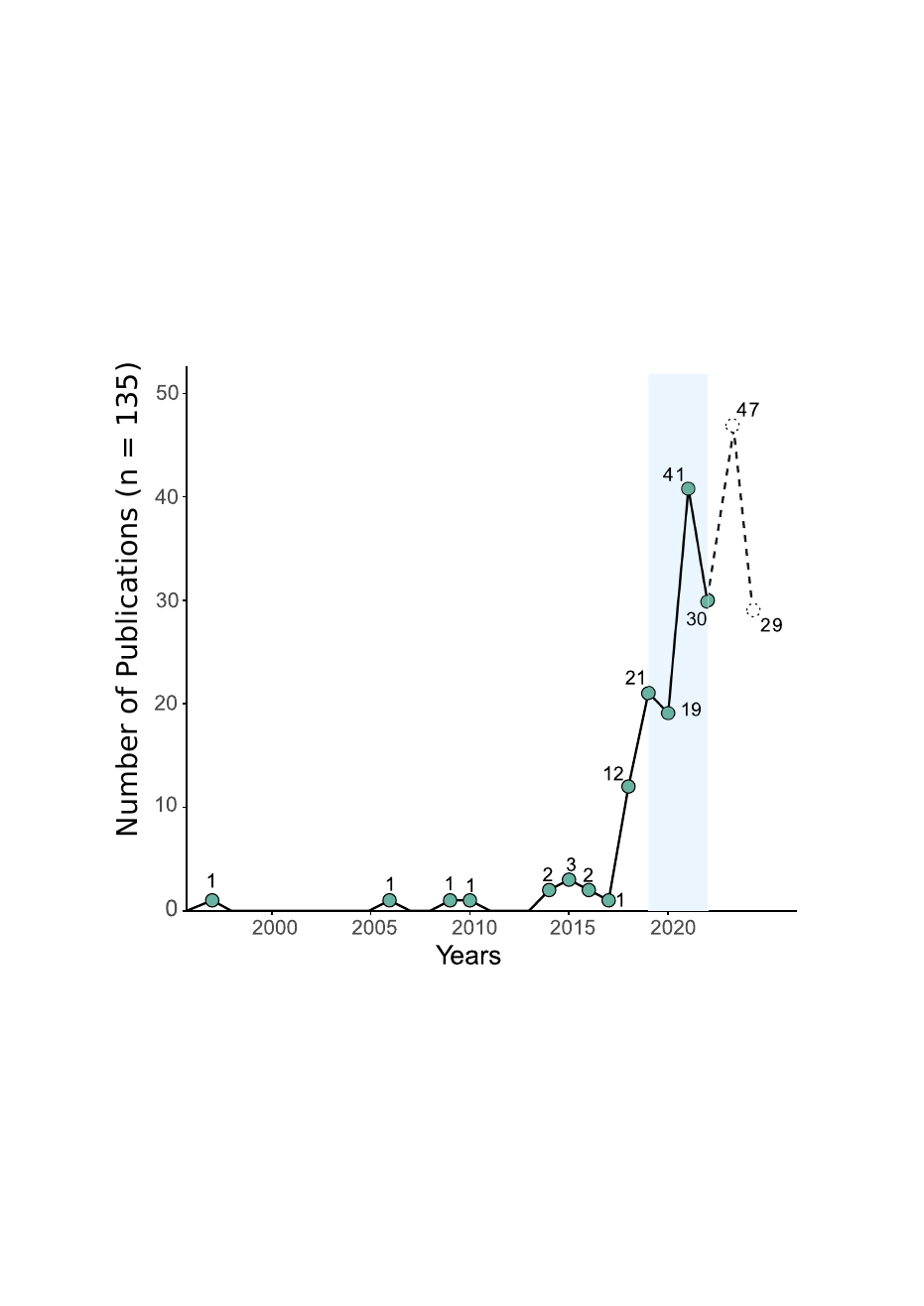}
\caption{Number of publications per year between 1997 and 2022, in light blue the articles published after 2018 concentrated more than 80\% of the publications. The dashed line represents publications from 1 January 2023 to 31 September 2024. Figure generated with \texttt{R 4.2.2} (code available in supplementary material 3) with additional editing in Inkscape.}
	\label{fig3}
\end{figure}

Through a chronological analysis of the reviewed articles [\textbf{\cref{fig3}}], we could distinguish a visible increase in the number of publications in recent years, with over 80\% articles reviewed published after 2018. From our quick search for publications between January 2023 and September 2024, which laid outside some of inclusion criteria, and thus we did not review, we obtained a total of 278 unique records. This number is slightly higher (by 12\%) than the number of records from 2021 to 2022 from the results of our automatic protocol search (n = 248). In comparison, the number of articles found that were published in 2021 and 2022 found from the automatic protocol search showed an increase of 82\% compared to those published in 2019 and 2020 (n = 136). A total of 76 records from the 278 obtained from our 2023 and 2024 search passed the automatic screening protocol. Although publications from the last trimester of 2024 were not included, our results suggest a stabilisation in terms of publication numbers in contrast to the large increase after 2020.

Focusing now only on the 135 publications reviewed, we observed that 10 journals had published nearly half of them (67 records, 49.63\%), from a total of 68 different journals observed. Out of these, 6 journals had published nearly a quarter of all articles reviewed [23.70\%, \textbf{\cref{tab3}}]. The journal \textit{Remote Sensing} had the most publications, followed by the \textit{Journal of Archaeological Science}. However, if we use the journal’s impact factor, as well as the journal-wide \textit{h}-index, multiplying them by the number of publications, we obtain two rudimentary research impact metrics across all the publications reviewed. From this analysis, \textit{PLOS One} obtained the highest combined score for the \textit{h}-index-based metric and second for the impact factor–based metric (IF). An analysis of the publication policies revealed that a total of 95 articles (70.37\%) were published with an open access modality (including gold, silver, or bronze) or were otherwise freely accessible. Some bias towards larger journals or more open access articles may have been occurred due to our search strategy (which for example, did not make use of Scopus) or search queries [see boxes \textbf{\hyperref[box:1]{1}} and \textbf{\hyperref[box:2]{2}}]. Additional bias could have come from a lower interest from these journals in publishing machine learning applications compared to more well-established approaches.

\begin{table}[ht]
    \centering
    \begin{tabular}{cccccc}
    \toprule
        \textbf{Journal} & \textbf{Num. of Articles} & \textbf{\textit{h}-index} & \textbf{n. \textit{h}-index} & \textbf{IF} & \textbf{n. IF} \\
        \midrule
        Remote Sensing & 15 & 193 & 2895 & 4.2 & 63 \\ 
        Journal of Archaeological Science & 14 & 152 & 2128 & 2.6 & 36.4 \\
        PLOS One & 10 & 435 & 4350 & 3.75 & 37.5 \\ 
        Scientific Reports & 6 & 315 & 1890 & 3.8 & 22.8 \\ 
        Journal of Computer Applications in Archaeology & 5 & 15 & 75 & N/A & N/A \\ 
        Archaeological Prospections & 4 & 46 & 184 & 2.1 & 8.4 \\ 
        Journal on Computing and Cultural Heritage & 3 & 35 & 105 & 2.7 & 8.1 \\ 
        Archaeological and Anthropological Sciences & 3 & 42 & 126 & 2.14 & 6.42 \\ 
        Palaeogeography Palaeoclimatology Palaeoecology & 3 & 177 & 531 & 2.6 & 7.8 \\
        Virtual Archaeology Review & 3 & 17 & 51 & 1.6 & 4.8 \\ 
        \bottomrule        
    \end{tabular}
     \vspace{0.25cm}
        \caption{The ten most represented journals and their \textit{h}-index and Impact factor (IF) score and total score by the number of articles, n = 135. Metrics were consulted on 14/07/2024 on the paper website for the impact factor or on SJR for the h-index [\textbf{\hyperref[annexe]{supplementary file}}].}
	\label{tab3}
\end{table}

\begin{figure}[t]
	\centering
	\includegraphics[width=1\linewidth, trim = {0cm, 0cm, 0cm, 0cm}, clip]{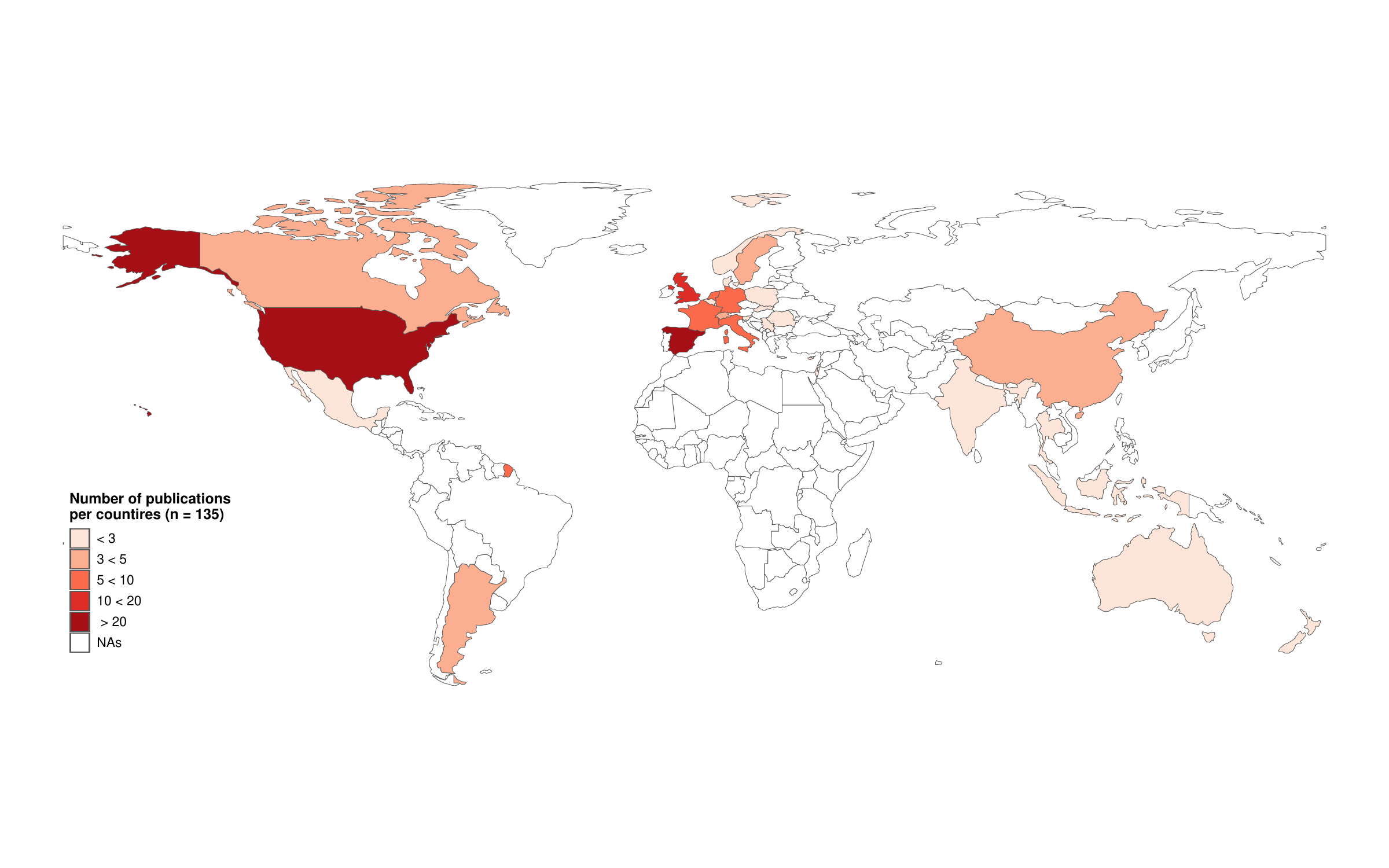}
\caption{Number of articles published per country based on the country of the first author’s affiliation. Figure generated with \texttt{R 4.2.2} (code available in supplementary material 3).}
	\label{fig4}
\end{figure}

The geographical distribution of the main institutions of the first authors from our set of reviewed articles shows that European and Anglosphere countries, sometimes referred to as the “Global North”, are over-represented [\textbf{\cref{fig4}}], already observed by Davis [\citeyear{davis_geographic_2020}]. Although one could also draw a line between the northern and southern hemispheres to divide the geographical distribution of the analysed publications, this division is not as clear, as countries such as New Zealand and Australia, as well as to some extent Argentina, which is part of the “Global South”, challenge this notion.

\subsection{Review findings}
\subsubsection{Statistical overview}
\begin{figure}[p]
	\centering
	\includegraphics[width=0.95\linewidth, trim = {0cm, 0cm, 0.1cm, 0.1cm}, clip]{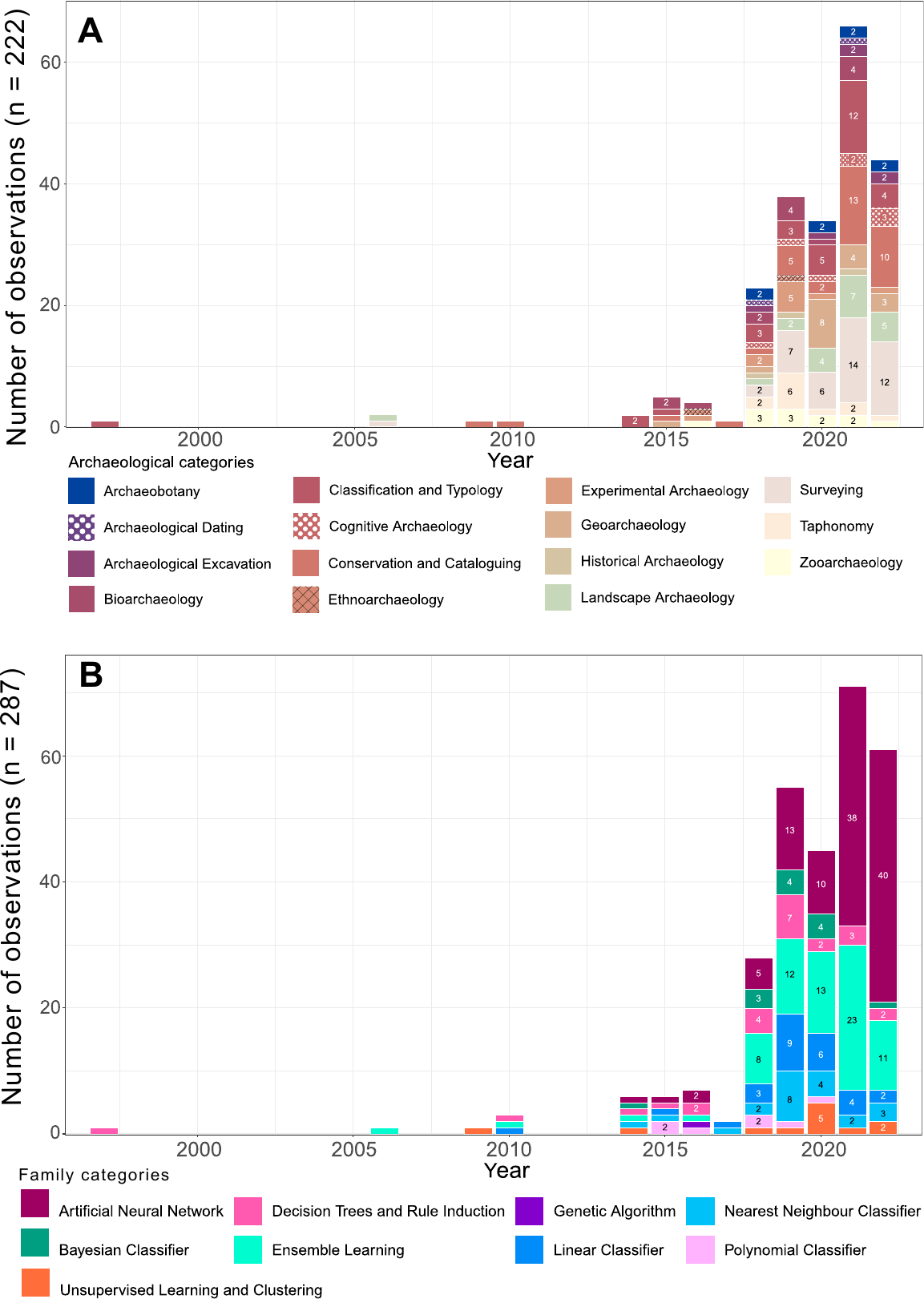}
\caption{(A) Number of articles from each archaeological subfield between 1997 and 2023. (B) Number of articles from each architecture class between 1997 and 2023. Empty bar charts represent the number 1. Figure generated with \texttt{R 4.2.2} (code available in supplementary material 3).}
	\label{fig5}
\end{figure}

Among the different subfields of archaeology present in our review, surveying (and site prospection), conservation and cataloguing, as well as classification and typology, are the most represented in our dataset and together account for 49\% of all study cases [\textbf{\hyperref[fig5]{Fig. 5A}}]. Landscape archaeology and geoarchaeology are also well represented with 20 and 17 attributed study cases respectively. The oldest applications (1997 - 2016) of machine learning in archaeology are linked with the subfields of classification and typology as well as conservation and cataloguing. In recent years, there has been a general increase in all topics with no clear trend visible, except for the subfields of surveying as well as conservation and cataloguing, which have considerably increased since 2021 [\textbf{\hyperref[fig5]{Fig. 5A}}]. Only the subfields of zooarchaeology, archaeobotany, and archaeological excavation remain nearly constant through time.

\begin{figure}[t]
	\centering
	\includegraphics[width=1\linewidth, trim = {0.25cm, 0.25cm, 0.25cm, 0.25cm}, clip]{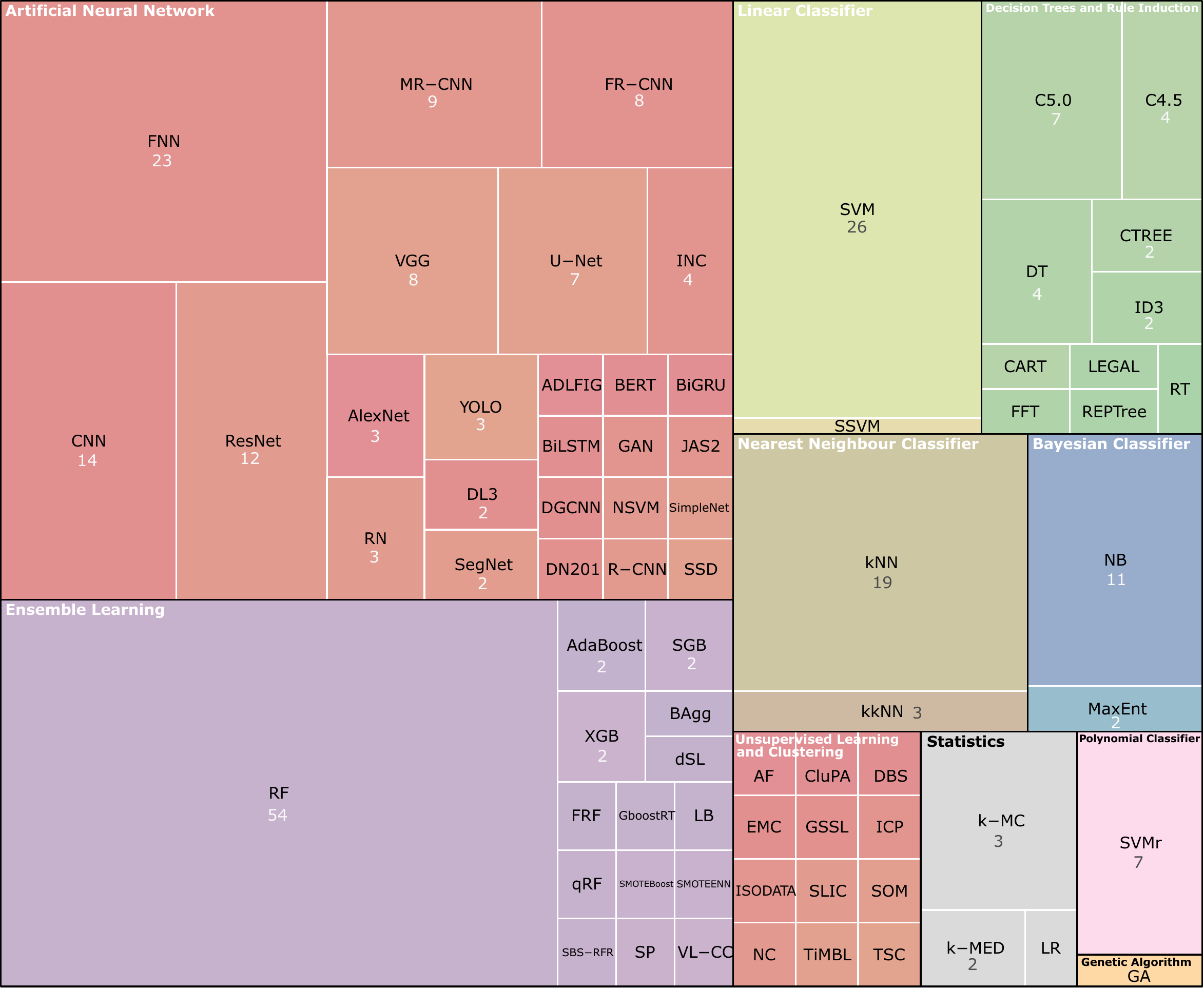}
\caption{Tree map of the different models seen in our corpus as well as the family of models they belonged to in our categorisation. The models with no number label have been counted only once. Figure generated with \texttt{R 4.2.2} (code available in supplementary material 3).}
	\label{fig6}
\end{figure}

Examining the different architecture of machine learning methods used, we observed that artificial neural networks (ANNs) and ensemble learning represent 62\% of all study cases, with 110 and 70 occurrences respectivly [\textbf{\hyperref[fig2]{Figs. 2} and \hyperref[fig5]{5B}}]. Linear classifiers, decision trees, rule induction, and nearest neighbour classifiers were also well-represented, accounting for 25\% of total applications. An increase of the use of ANNs in our list of reviewed articles from 2021 onwards could be observed, with the number of applications multiplying four-fold compared to the number of applications in the years 2019 and 2020 [\textbf{\hyperref[fig5]{Fig. 5B}}]. At a more granular scale, we found a total of 70 machine learning models [\textbf{\hyperref[fig6]{Fig. 6} and \hyperref[annexe1]{Annexe 1}}]. In addition, we found a particularly high diversity in the unsupervised learning and clustering category of model families, with a unique model for each of the twelve recorded applications for the category [\textbf{\hyperref[fig6]{Fig. 6} and \hyperref[annexe1]{Annexe 1}}]. Despite the dominance of ANNs [\textbf{\hyperref[fig6]{Fig. 6}}], random forest was the most common individual model (n = 54). We found that across all  articles reviewed, the mean number of models used per study case was 2.12, with a high disparity across different publications, with some applying only one model, while articles such as Bataille et al. [\citeyear{bataille_bioavailable_2018}] and Courtenay et al. [\citeyear{courtenay_combining_2019}] tested six different models at once. Classification was the most common form of data evaluation, applied in more than 75\% of study cases (n = 112). On the other hand, only 19\% (n = 29) were regression models, and clustering was only represented with six uses [6\%, \textbf{\cref{fig2,fig8,fig9,fig10}}].

\par It was more difficult to observe patterns for the other features we collected, as the attribution to one or several categories for an article was more subjective to the person reviewing. Two of the \textit{a posteriori} study tasks are highly prominent: automatic structure detection and artefact classification. These two tasks account for 45\% (n = 67) of all study case tasks [\textbf{\cref{fig2}}]. Tasks such as taphonomic classification, archaeological predictive models, and architectural element classification are well represented with 5 to 13 study cases each, whilst nine task categories (\textit{e.g.} sourcing, species classification, movement recognition) were only relevant for around 5 study cases.

\par
\begin{figure}[t]
	\centering
	\includegraphics[width=1\linewidth, trim = {0cm, 0cm, 0cm, 0cm}, clip]{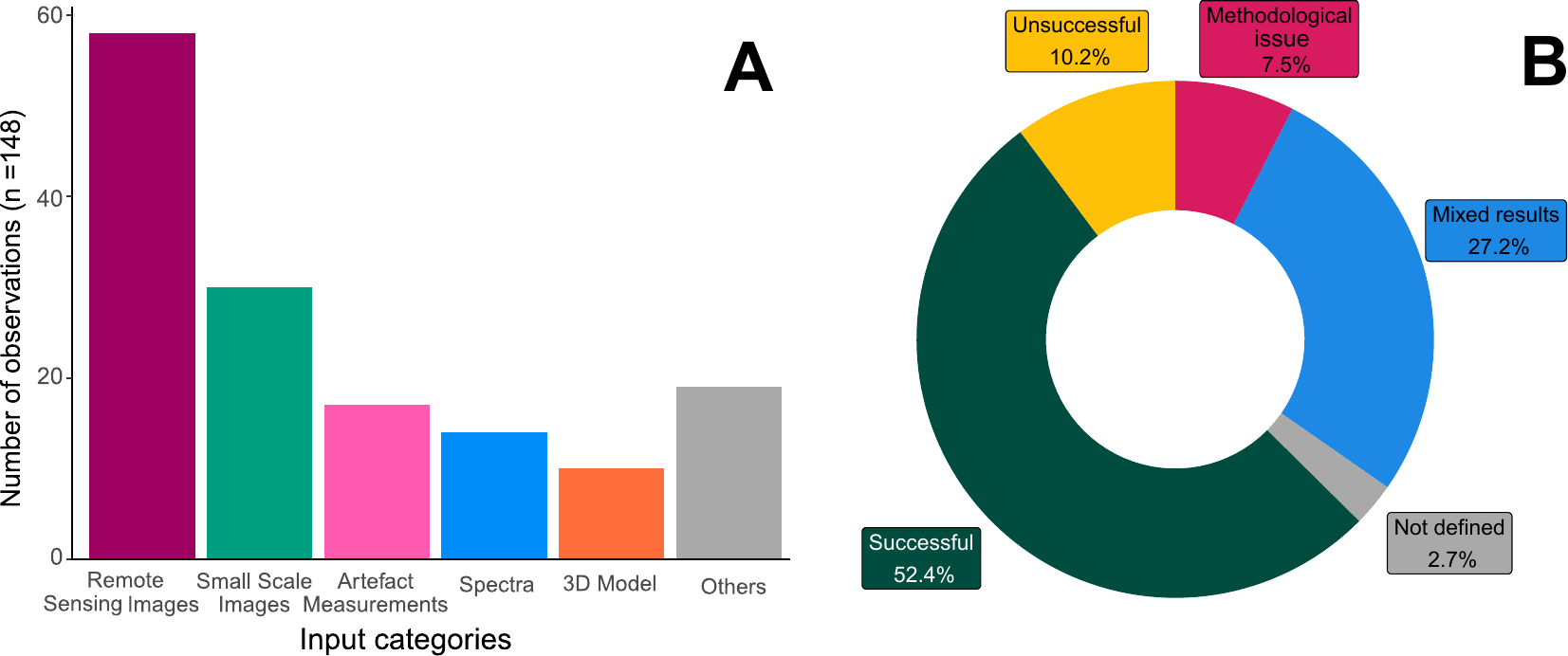}
\caption{(A) The five more represent classes of input data among the reviewed papers, n = 148. (B) Results of the reviewed papers according to the authors or presented results, n =  147. Figure generated with \texttt{R 4.2.2} (code available in supplementary material 3).}
	\label{fig7}
\end{figure}

 For the different types of input data observed in all study cases we reviewed, remote sensing images were the most represented, with around 40\% (n = 58) of the total. Four other input types (small-scale images, artefact measurements, spectra, and 3D models) were also well represented [\textbf{\hyperref[fig7]{Fig. 7A}}], being present in between 6 and 16\% of all study cases reviewed. The remaining nine input types were poorly represented, being present in only 14\% of all study cases.

\par Finally, according to the assessment of article authors or (if not discussed) the results metrics, we could observe that most applications (n = 79) had reported “successful” outcomes [\textbf{\hyperref[fig7]{Fig. 7B}}], with an additional 40 mixed results (27.2\%). Only 15 applications (10.2\%) were reported to have been “unsuccessful”, whilst we noted 9 (6.1\%) study cases whose application had important methodological issues, and classified them as such even if the authors reported a successful outcome. Finally, four (2.7\%) study cases did not have a defined outcome. These results, however, could be affected by the well-reported issue of publication bias against the reporting of negative results \citep{song_publication_2013}, as well as the related file-drawer effect \citep{rosenthal_file_1979,sparks_prying_2009}, which can strongly affect the perception and production of scientific results \citep{song_dissemination_2010}. 

\subsubsection{Aggregated information}

Through the aggregation of the data annotated for the study tasks, method families, archaeological subfields, and input types we obtained an overview of the relationship between the objectives of the study cases and the means used to access them.\\
Correlations between tasks and model families are highly heterogeneous [\textbf{\cref{fig8}}]. Whilst the highest correspondence rate between taphonomic classification and unsupervised learning and clustering methods is 26\%, the correspondence between artefact classification and ANNs is much higher, at 49\%. Even stronger correlation rates could be observed between archaeological predictive models and ensemble learning methods at 66  and between automatic structure detection and ANNs, at 60\%.\\
The correlations between archaeological subfields and study tasks are lower than the correlations visible for architecture [\textbf{\cref{fig9}}]. Articles dealing with environmental reconstruction had their highest correlation at only 33  with the subfield of geoarchaeology. Studies treating artefact classification problematics have a higher 45\% correspondence rate with subfields of classification and typology. The highest correlation rate, however, is present between the task of automatic structure detection and the subfield of surveying, at 80\% correlation.

\par Finally, and perhaps expectedly, several of the most represented tasks had the largest number of study cases with reported unsuccessful or mixed results [\textbf{\cref{fig10}}]. Studies dealing with automatic structure detection (25.85\% of total) and artefact classification (19.73\% of total) accounted for 60\% and 33\% (respectively) of studies that reported mixed or unsuccessful results. Other less represented tasks have higher success rates. Articles dealing with artefact prediction are successful in 80\% of cases, whilst articles dealing with  environmental reconstruction all report successful results.

\subsubsection{Classification of artefacts and animal remains}
\begin{figure}[b]
	\centering
	\includegraphics[width=1\linewidth, trim = {0cm, 0cm, 0cm, 0cm}, clip]{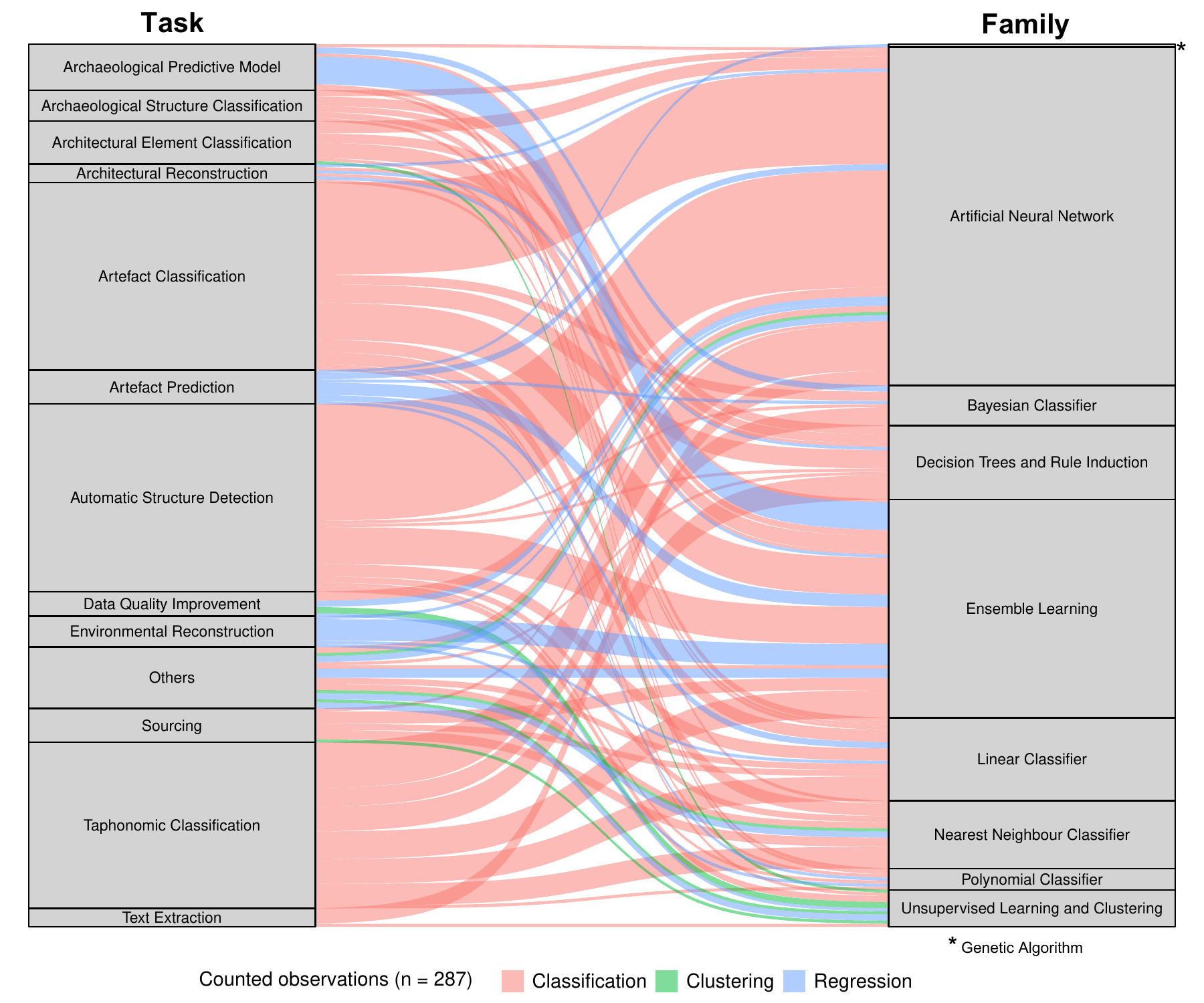}
\caption{Alluvial diagram of the different tasks in the analysed studies on the left, the related architecture of machine learning models on the right and the evaluation process in the background. Tasks and architectures poorly represented (n < 5) have been classified as “others”. A study might have applied numerous models or its research objectives could be classified into more than one task. In such cases, we created multiple entries for each paper where applicable [\textbf{\hyperref[annexe]{supplementary file}}]. Figure generated with \texttt{R 4.2.2} (code available in supplementary material 3).}
	\label{fig8}
\end{figure}

\begin{figure}[b]
	\centering
	\includegraphics[width=1\linewidth, trim = {0cm, 0cm, 0cm, 0cm}, clip]{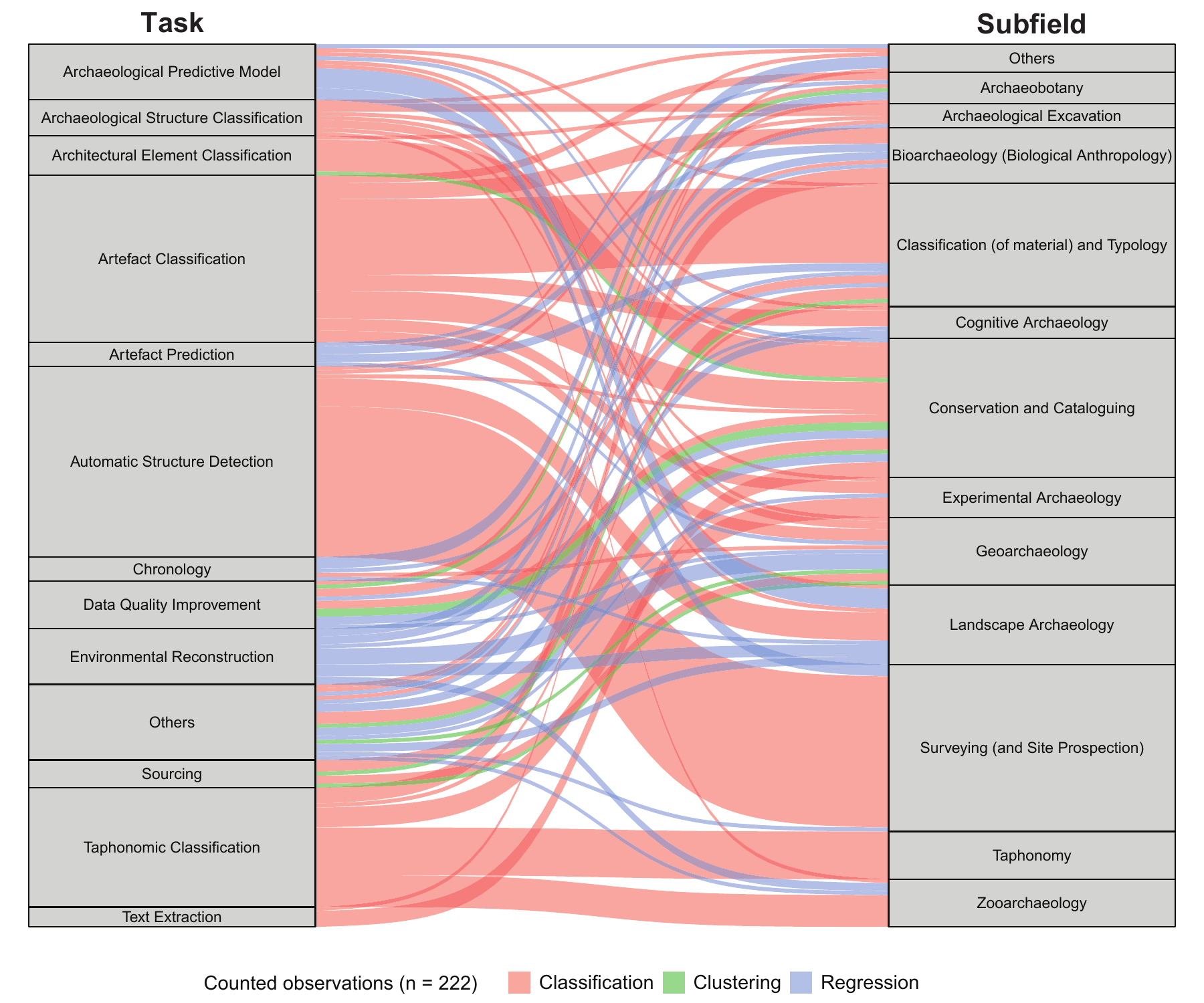}
\caption{Alluvial diagram of the different tasks in the analysed studies on the left, the related archaeological subfields on the right with the evaluation process in the background. Tasks and subfields poorly represented (n < 5) have been classified as “others”. A study might have been attributed to several subfields or its research objectives could be classified into more than one task. In such cases, we created multiple entries for each paper where applicable [\textbf{\hyperref[annexe]{supplementary file}}]. Figure generated with \texttt{R 4.2.2} (code available in supplementary material 3).}
	\label{fig9}
\end{figure}

Whilst classification tasks were mainly performed on ceramic material (31\%), as in Hörr et al. [\citeyear{horr_machine_2014}] or Anichini et al. [\citeyear{anichini_automatic_2021}], a wide range of different materials were used in our dataset of reviewed articles. Studies on coins \citep{boon_digital_2009}, stone tools \citep{macleod_quantitative_2018,pargeter_understanding_2019,emmitt_machine_2022}, archaeobotanical seed remains \citep{landa_accurate_2021}, phytoliths \citep{berganzo-besga_automated_2022}, ivory figurines \citep{gansell_stylistic_2014}, were observed. Moreover, 17\% of study cases were conducted on cave art images, such as in Kogou et al. [\citeyear{kogou_remote_2020}], or Horn et al. [\citeyear{horn_artificial_2022,horn_boat_2022}]. On all classification study cases 17\% (n = 5) dealt with the bioarchaeological classification of human or animal remains. Due to the variety of materials studied, the input data for models were very diverse. Small-scale images accounted for 48\% (n = 14) of all input data for classification task, metric data for  27\% (n = 8), as well as less frequent input data types such as multilayer images of artefact (\textit{e.g.} depth maps, 3D model layers), spectra and remote sensing images all together accounted for 23\% (n = 7).\\
In our corpus, two machine learning method families emerge as the foremost choices for artefact classification, ANNs represent 50\% of the models used, followed by ensemble learning, accounting for 31\% of the models [\textbf{\cref{fig8}}]. Bayesian classifiers, as well as decision tree and rule induction are also present, but represent less than 15\% of the model architectures for the artefact and animal remains classification task.

\subsubsection{Archaeological predictive models (APMs)}
\label{3.3.4}
Ten studies in our review focused on archaeological predictive models (APMs), dealing with either human landscape occupation patterns, or human-environment interactions. Input data were in 77\% (n = 7) of all study case larger-scale raster images from a diverse range of natural covariates or remote sensing images.\\
The most common method family applied was ensemble learning, representing 62.5\% (n=10) of the entries for this task. A unique model entry exclusively applied to this task was Maximum Entropy (\textit{MaxEnt}) \citep{benner_combining_2019, yaworsky_advancing_2020}. Regarding the results from APMs, 66\% (n=6) were classified as successful, with the remaining 44\% (n=4) classified as partially successful or unsuccessful, such as in Hansen and Nebel [\citeyear{hansen_prioritizing_2020}] or Miera et al. [\citeyear{miera_large-scale_2022}].

\subsubsection{Automatic structures detection}
Representing 25\% (n = 38) of all study cases, automatic structures detection, also known as geographical object-based image analysis (GEOBIA), is the most prominent task category in our review. Input data came from a wide range of remote sensing images, mainly LiDAR or airborne laser scanning (ALS), used in 50\% of all study cases in this task, or various satellite image collections (23\%), such as in Menze et al. [\citeyear{menze_detection_2006}] or Orengo et al. [\citeyear{orengo_automated_2020}]. The remaining 27\%  input data used were either UAV ortho-images, such as in Orengo and Garcia-Molsosa [\citeyear{orengo_brave_2019}], Monna et al. [\citeyear{monna_machine_2020}], Agapiou et al. [\citeyear{agapiou_detection_2021}], Altaweel et al. [\citeyear{altaweel_automated_2022}], and Fisher et al. [\citeyear{fisher_multidisciplinary_2022}], historical maps, such as Garcia-Molsosa et al. [\citeyear{garcia-molsosa_potential_2021}], and since 2020, ground-penetrating radar (GPR), or even bathymetric data \citep{febriawan_detection_2020, bordon_automatic_2021}.\\
ANNs were the most prominent family of methods for automatic structure detection, accounting for 62\% of all models applied for this task. Different ANN models were used, the most popular being mask region-based convolutional neural networks (MR-CNN), which accounted for 13\% of all models applied for automatic structure detection, followed by faster region-based convolutional neural networks (FR-CNN), and U-Net, which each accounted for 8\% and 7\% of the models used. However, the most popular model was random forest, which accounted for 16\% of all models used for this task. The results of applications in this task were mainly reported as unsuccessful or partially successful (59\%, n = 31).

\subsubsection{Digital heritage}
Articles focused on this task were mainly represented by those dealing with the classification and reconstruction of architectural elements, with a total of twelve study cases (8\% of all 147 study cases reviewed). The models in this task category  mainly used small-scale images from architectural photos as input (58\%, n = 7), but also point clouds and 3D models in 33\% of the revived study cases, such as in Grilli and Remondino [\citeyear{grilli_classification_2019}] or in Matrone and Martini [\citeyear{matrone_transfer_2021}].\\
The models applied in this task are somewhat equally divided between two families of models: ensemble learning with 35\% of all study cases, and decision trees and rule induction models accounting for 25\% of total models applied. Machine learning applications for this task  were generally negatively assessed, in comparison to applications for other tasks, with only 40\% of study cases reporting successful results.

\subsubsection{Text analysis}
We counted five study cases that applied machine learning methods to text analysis tasks (3\% of all 147 study cases), and all except one \citep{dhivya_tamizhi_2021} used text as input data for their model. Almost all applied models were based on ANNs (83\%, n = 5), such as in Dhivya and Devi [\citeyear{dhivya_tamizhi_2021}] and Brandsen and Lippok [\citeyear{brandsen_burning_2021}], only Boon et al. [\citeyear{boon_digital_2009}] used a memory-based learning (MBL) model, part of the category of unsupervised learning and clustering algorithms. The results of the applications for this task were overwhelmingly negatively assessed, with four out of the five studies reporting mixed or unsuccessful results.

\subsubsection{Taphonomic classification}
\label{3.3.8}
\begin{figure}[b]
	\centering
	\includegraphics[width=1\linewidth, trim = {0cm, 0cm, 0cm, 0cm}, clip]{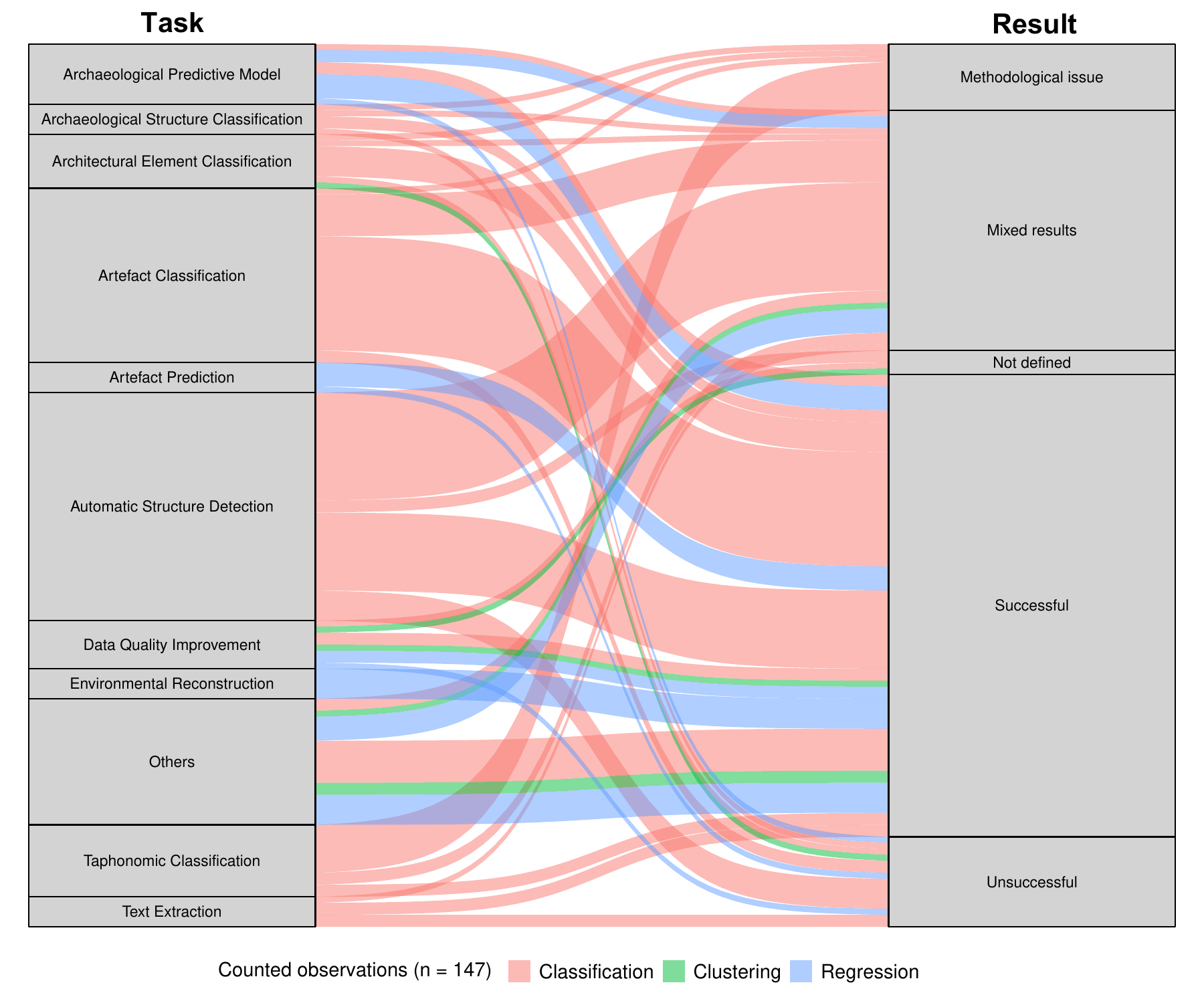}
\caption{Alluvial diagram of the different tasks in the analysed studies on the left, the related results on the right with the evaluation process in the background. Tasks poorly represented (n < 5) have been classified as “others”. A study might have its research objectives classified into more than one task. In such cases, we created multiple entries for each paper where applicable [\textbf{\hyperref[annexe]{supplementary file}}]. Figure generated with \texttt{R 4.2.2} (code available in supplementary material 3).}
	\label{fig10}
\end{figure}

We reviewed a total of twelve study cases categorised as dealing with the classification of taphonomic features on archaeological finds (8\% of all case studies). All of these articles were applied to osseous material, and all but two of them deal with bone surface modifications (BSMs), with the remaining two dealing with bone breakage. Bone surface modifications are marks left on the surface of bones by agents such as hominin butchery (\textit{e.g.} using stone tools), animal carnivory, or trampling. The goal of all the articles under discussion was to uncover what agent caused the specific taphonomic process (BSM or breakage). Input data were primarily metric and categorical variables of bone marks or breaks in six cases (50\%), or microscope images in three cases (25\%), with the remaining three (25\%) using PCA scores from the analysis of geometric morphometric landmarks.\\
Many of the articles reviewed applied a large number of machine learning methods as well as methods that we do not consider machine learning [such as PLSDA and MDA; \textit{e.g.} \citealt{dominguez-rodrigo_successful_2018, abellan_high-accuracy_2022}] to compare their performance on the specific task, whilst others [\textit{e.g.} \citealt{byeon_automated_2019, cifuentes-alcobendas_deep_2019}] were more restrained. This meant that there was no clear machine learning method or family of methods that was used much more frequently than others, although neural networks were applied in all but one of the articles reviewed (see “Results”), which is not the case for other methods also commonly used in these articles, such as SVMs or random forests.\\
ANNs were applied in all the reviewed study cases except for one \citep{aramendi_who_2019} and represent 28.3\% (n = 15) of all models applied for taphonomic classification. Ensemble learning, linear classifiers, decision trees and rule induction, nearest neighbour classifiers, and Bayesian classifiers all represented between 17\% and 11\% of the referenced architectures for this task category.\\
However, many of the articles found were deemed through our reviewing process to have had important methodological issues, especially due to under-reporting or under-detailing of the specific methodology used [\textbf{\cref{fig10}}, cf. \ref{4.2.6}].

\section{Discussion}
\subsection{General considerations on the corpus}
\label{4.1}
A possible initial observation is the comparatively late interest in machine learning applications in the field of archaeology compared to other sciences \citep{dramsch_70_2020,padarian_machine_2020,shehab_machine_2022,wang_review_2022}. Indeed, the exponential increase in publications related to machine learning applications in archaeology can be dated from the period between 2018 and 2019 onwards [\textbf{\cref{fig3}}]. This late development could arguably be attributed to a tradition of suspicion in archaeology \citep{yoffee_archeologie_2010}. Researchers sometimes consider archaeological data as scattered, complex \citep[p.7]{smith_approaches_2011} and unable to be explained by models that are not, fully, human-made. The uncertainty for selecting the best model to analyse the complexity of past human action is not recent \citep{doran_computer_1990} but has been exacerbated with machine learning and even more with deep learning models, often opaque, \textit{“black boxes”} in their internal logic \citep{ramazzotti_modeling_2020,fiorucci_machine_2020,bickler_machine_2021,calder_use_2022}. Furthermore, it appears that some sub-disciplines of archaeology have until 2022 yet to exploit the new development of machine learning to the same degree as other sub-disciplines. Following our classification of archaeological subfields (see supplementary material), ethnoarchaeology, archaeobotany, and cognitive archaeology are under-represented in our corpus [\textbf{Figs. \hyperref[fig2]{2} \hyperref[fig5]{5B}}]. Although the reasons for such an underrepresentation remain unclear, it is likely that many research questions in these subfields are less suitable for machine learning applications at this current time, or at least to lesser degree than other sub-disciplines \citep{marom_current_2025}.

\par Regarding publication policies across our list of reviewed articles, the high representation of open-access papers in a field where the traditional closed-access model of publications is widespread [see below; \citealt{marwick_standard_2018}] could be a signal for a paradigm change in which scholars seek to share their knowledge more broadly and fairly \citep{marwick_open_2017,nicholson_will_2023}. While many journals that deal specifically with computer science applications in archaeology (\textit{e.g. Archaeologia e Calculatori, Internet Archaeology}, or \textit{Journal of Computer Applications in Archaeology}) only have open access publication routes, more classical and broader journals like the \textit{Journal of Archaeological Research, Journal of Archaeological Science} or \textit{Antiquity} provide mainly traditional non-open access papers, with open access papers comprising only (respectively) 10\%, less than 5\% and less than 2\% of all articles on the 1st of September 2024. These results are unlike observations made for articles dealing with machine learning applications in other disciplines such as soil science \citep{padarian_machine_2020}, where paid-access routes vastly dominate that field of research. However, researchers have to keep in mind the existence of non-open access journal articles and book chapters to avoid a FUTON bias \citep{wentz_visibility_2002}, which will lead to not exploiting part of the wider research results in the field. This points to a wider adoption of both FAIR (Findable, Accessible, Interoperable, Reusable) data principles \citep{wilkinson_fair_2016, nicholson_will_2023} and CARE (Collective benefit, Authority to control, Responsibility, Ethics) principles \citep{carroll_care_2020, gupta_care_2023}. Nonetheless, if the degree of open access of machine learning applications in archaeology seems promising, the large disparity borne from colonialism remains unfortunately wide \citep{davis_geographic_2020}. The difference in the number of publications from “Global North” and “Global South” countries can be directly linked with the gap in the economic development between higher and lower-income countries, leading to the underdevelopment of science and technology \citep{sagasti_underdevelopment_1973, allik_factors_2020}. Increasing collaborations between different countries and institutions could fill this gap between higher and lower-income countries \citep{sonnenwald_scientific_2007}.

\subsection{How is machine learning used across archaeology?}
In the earlier stage of machine learning applications in archaeology, many archaeological questions could only be answered by seeking experts in informatics, mathematics, and physics \citep{menze_detection_2006,boon_digital_2009,menze_mapping_2012,horr_machine_2014}. However, a noticeable shift has occurred over time. Archaeologists have progressively embraced programming their own tools (by using \textit{e.g.} \texttt{Python, R, Matlab, Java}), and have also turned to specialised software solutions such as Grilli and Remondino [\citeyear{grilli_classification_2019}], Garcia-Malsosa et al. [\citeyear{garcia-molsosa_potential_2021}], Demjam et al. [\citeyear{demjan_laser-aided_2022}] or Casini et al. [\citeyear{casini_humanai_2023}]. Nowadays, in many studies, archaeologists are both asking and answering the questions themselves \citep{orengo_brave_2019,orengo_automated_2020,yaworsky_advancing_2020,carter_when_2021}. The CAA Special Interest Group on Scientific Scripting Languages in Archaeology (\href{https://sslarch.github.io/}{\textit{SIG-SSLA}} ) is one example of this polyvalency of archaeologists developing scripting or machine learning tools for their problems. In certain instances, computer science specialists independently pose questions and generate tailored answers for specific applications \citep{toler-franklin_multi-feature_2010,ushizima_materials_2020,wunderlich_hyperbola_2022}. However, there is always a question of the ratio between the time spent developing new skills \citep[pp. 109-112]{aldenderfer_quantitative_1998} and the increasing complexity, if not in programming, then in understanding, of machine learning methods, making it more and more difficult for an archaeologist to master all abilities needed to develop an entire workflow for a computer-based approach.

\par The understanding between the one who poses the question and the one who answers is crucial, as the nature of the answers obtained depends intricately on the questions asked. Therefore, archaeologists should be careful to verify the suitability and validity of the applied methods, as only they can understand the full complexity of what happens outside these \textit{black boxes}. Orton once stated that: “... \textit{the mathematician’s job is not to tell the archaeologist what to do – the archaeologist maintains his responsibility for what he does - but to help him decide how to do it.}” \citep[pp. 15-16]{orton_mathematics_1980}. This remains easily applicable to other specialists such as informaticians, especially relevant here, to this day. However, defining a single, universally applicable methodology for archaeological applications of machine learning methods, already a very large and diverse group, or other quantitative methods remains challenging (\textit{e.g. “Statistical Cycle”}, \citealt[p. 20, fig. 1.3]{orton_mathematics_1980}), a pragmatic approach can highlight several best practices depending on the desired task [\textbf{\cref{fig11}}].

\subsubsection{Classification of artefacts and animal remains}
\label{4.2.1}
Classifying artefacts and animal (including human) remains into categories is a central aspect of archaeological work \citep{read_archaeological_2018}. Archaeologists have looked for solutions in computer science since the second half of the twentieth century [cf. \ref{1} \citealt{binford_preliminary_1966}, \citealt[pp. 156-178]{orton_mathematics_1980}]. It is therefore not surprising to see machine learning applications developed for this purpose already very early [\textbf{\hyperref[fig5]{Fig. 5A}}]. Earlier papers generally applied methods such as decision trees, ensemble learning, Bayesian classifiers, as well as linear classifiers combined with nearest neighbour classifiers \citep{nguifo_plata_1997,horr_machine_2014,mircea_sex_2015,macleod_quantitative_2018,canul-ku_classification_2019,pargeter_understanding_2019,gonzalez-molina_distinguishing_2020}. Since 2020, however, ANNs have gained more traction [\textit{e.g.} \citealt{ushizima_materials_2020,graham_towards_2020}], and now represent the most commonly used family of methods applied for this task [\textbf{\cref{fig8}}, \citealt{ling_findings_2024}]. ANNs, especially CNNs, are standard for computer vision tasks like object classification \citep{guo_deep_2016,chai_deep_2021}. Dimensionality reduction methods (\textit{e.g.} PCA and MDA) are often performed alongside machine learning algorithms, and can be an effective way to improve model accuracy when applied correctly [\textit{e.g.} \citealt{muzzall_novel_2021}]. However, their ability to distinguish categories of artefacts seems limited compared to machine learning methods such as random forest \citep{cole_evaluating_2022}.

\par Concerns about the validity of certain forms of typological classification, both human-made and computer-based, as a practice have been previously raised, especially from North American researchers \citep{adams_typological_1991,horr_machine_2014}. Other researchers, however, do not hesitate in using computational approaches for classification \citep{pargeter_understanding_2019}, and even creating new classification methodologies with the help of machine learning. Notably, Hörr and others, seeking a less subjective approach to typological classification, developed a three-layer method for this purpose [\citealt[pp. 136-137]{horr_algorithmen_2011}, \citealt{horr_machine_2014}], successively incorporating unsupervised, semi-supervised, and supervised classification algorithms within a framework Fayyad et al. [\citeyear{fayyad_data_1996}] termed the \textit{Knowledge Discovery Process}. This approach is promising for the analysis of archaeological material, and could prove useful for many archaeological studies. Such an approach invites critical questions on the artificiality of typological classifications and opens new avenues for future studies such as those dealing with ceramic material or stone tools. We can also mention Martín-Perea et al. [\citeyear{martin-perea_application_2020}], who used a method similar to that of \citep{horr_machine_2014} with a three-level pipeline to detect fossiliferous levels of an archaeological or palaeontological site.

\par Assessing and comparing the outcomes of these diverse classifications is challenging, given the inherent variations in datasets, modelling techniques, and evaluation metrics employed across different studies. Nonetheless, various opinions have already emerged on the issue. Emmit et al. [\citeyear{emmitt_machine_2022}] and Jalandoni [\citeyear{jalandoni_use_2022}] are optimistic about using machine learning for the classification of archaeological material, though we could also add Hörr et al. [\citeyear{horr_machine_2014}], who are more cautious but remain hopeful for the future of the practice. Others are less enthusiastic, however, and point to mixed results \citep{demjan_laser-aided_2022,lyons_lidar_2022}. Overall it seems likely that machine learning applications for the classification of archaeological material will spread not only within the scientific community but also among professionals of archaeology, conservation institutions, and a wider public. The \textit{ArchAIDE} project is already a noteworthy example of an application of open-access machine learning tools for archaeology to serve the public at large \citep{gualandi_open_2021, anichini_automatic_2021, anichini_reflecting_2022}. Another example of a mobile application for ceramics classification is developed in Santos et al. [\citeyear{santos_automatic_2024}], and a broader project of automation and digitalisation tools for archaeological artefacts has been developed in the \textit{Automata} initiative \citep{naso_state_2025}. The reconstruction of ceramics based only on few pottery sherds has also been achieved with machine learning methods \citep{stamatopoulos_3d_2016, cardarelli_fragments_2024}. Another promising development has been made in Ruschioni et al. [\citeyear{ruschioni_supervised_2023}], where they classified ceramics based not on their typology but rather on their chemical properties, which is more data-compatible with a machine learning process.

\par In summary, as one of the core elements of archaeological work, the use of machine learning for the classification of artefacts has been debated theoretically and tested since the early 2000s. While ceramics are the most represented type of artefact used for machine learning-based classification in our corpus, a wide variety of materials have also been used (\textit{e.g.}, ivory, stone tools, etc.). With the majority (62\%, n = 18) of study cases using image data as inputs, the recent increase in performance of neural network models could spur further advances for artefact classification. The future of this discipline will also likely be linked with more data-driven material, standardised pre-processing, and user-friendly classification software, be it portable or not.

\subsubsection{Archaeological predictive models (APMs)}
Already present since the development of settlement patterns analysis \citep{willey_prehistoric_1953}, the tradition of APMs rose alongside the development of modelling and prediction theory in archaeology \citep{judge_quantifying_1988}. It is therefore surprising to see little interest in applying machine learning methods to this approach in recent years even down to 2022 [\textbf{\hyperref[fig5]{Fig. 5A}}]. APMs focus on predicting unknown or undiscovered sites so that they can be discovered and documented based on previous field observations and variable predictors, for example climate, topography and anthropic features \cite{brandt_experiment_1992, kvamme_there_2006}. Critics can point to the relative reluctance of researchers towards  APMs [\citealt[p.7, Table 1.1]{kvamme_there_2006}, \citealt[pp. 42-45]{lock_enhancing_2006}]. One main criticism about APMs is that “\textit{… predictive modelling as it is presently practiced is fundamentally about environmental determinism.}” \citep[p.19]{kohler_predictive_1988}. This statement is particularly true when we consider the strong influence of environmental and ecological studies in the field of landscape archaeology (cf. \ref{3.3.4}). However, even if this criticism has neither recent origins, nor is it confined to the past \citep{wheatley_making_2004, arponen_environmental_2019,kristiansen_who_2019}, responses to it exist. Coombes and Barber emphasise the interest of models despite their incorrectness, stating that ”\textit{A simple model cannot hope to replicate all the complexities of environment–culture relationships across a civilization, but one basic approach that can provide valuable insights is to treat human populations in ecological terms, with their ranges shifting in response to changing conditions}” \citep[p. 305]{coombes_environmental_2005}. Furthermore, new applications of APMs dedicated to cultural heritage preservation against looting \citep{el-hajj_interferometric_2021} can prove to be a promising solution to further aid in the protection of cultural heritage alongside more  intensive public funding.\\

\par The frequent use of the MaxEnt model for APMs, initially designed by ecologists \citep{phillips_maximum_2006, sillero_want_2021}, shows the influence of environmental niche concepts in archaeology \citep{demjan_laser-aided_2022,vernon_decomposing_2022, lundstrom_here_2024, yaworsky_effects_2024, yaworsky_neanderthal_2024}. MaxEnt is well suited to examining questions of dispersion and provides more robust and deterministic archaeological predictions due to it being a presence-only model, using background points rather than true absence or pseudo-absence data (used by other methods, including machine learning)  into its modelling: “\textit{As a result, MaxEnt is more suitable for archaeological data than the other predictive modelling approaches}” \citep[p. 14]{yaworsky_advancing_2020}. Furthermore, when compared to other models such as random forest, which can be prone to overfitting, MaxEnt appears as the best option for species dispersion problems \citep{valavi_predictive_2022}.

\par An essential consideration in APMs is the number of covariates, predictors, or variables required for the building of the models. In Yaworsky et al. [\citeyear{yaworsky_advancing_2020}], 55 predictors were used, Hansen and Nebel [\citeyear{hansen_prioritizing_2020}] employed 26 predictors, Friggens et al. [\citeyear{friggens_predicting_2021}] used 25 predictors, Castiello and Tonini [\citeyear{castiello_explorative_2021}] used 13 predictors, and Benner et al. [\citeyear{benner_combining_2019}] relied on 8 predictors. However, if data transformation techniques such as PCA are applied, which seek to reduce the number of covariates whilst maximising the impact of those that remain, a smaller number of predictors can be used for the final model instead \citep{hansen_prioritizing_2020,yaworsky_advancing_2020,el-hajj_interferometric_2021}. Research on species distribution models assess the maximum number of predictors at $k = \frac{(n-50)}{8}$ \citep{field_discovering_2012}, $k$ being the number of predictors and $n$ the number of occurrences of the species.

\par The large number of variables required, the complex preprocessing steps needed, and the difficulty in selecting and interpreting the influence of the model variables, make APMs a time-consuming and less attractive method than approaches based on image recognition, which can be simpler to apply. Furthermore, the theoretical criticisms discussed above must also be added to the methodological difficulty of applying APMs, which could lead to a desire to avoid further complexity in the methods in the form of machine learning methods, which could explain their low numbers compared to applications for other tasks.

\subsubsection{Automatic structures detection}
\label{4.2.4}
Over the past decade, there has been an explosion in machine learning methods applied to automatic structure detection tasks [\textbf{\hyperref[fig5]{Fig. 5A}}]. Though earlier applications were mostly based on satellite STRM or ASTER images \citep{menze_detection_2006,menze_mapping_2012}, the recent explosion of light detection and ranging (LiDAR) and airborne laser scanning (ALS) data \citep{wurzer_agent-based_2015,gillings_archaeological_2020}, has led to an increase in interest for this field.  While earlier studies employed random forest methods \citep{menze_detection_2006,guyot_detecting_2018,stott_searching_2019}, recent papers mainly applied ANNs, which aligns with the predominance of image data and neural network architectures that excel at image classification tasks. The recent multi-scale region-based convolutional neural network (MR-CNN) model, adapted from convolutional neural networks (CNNs), has been eagerly applied by archaeologists, as its \textit{raison d’être} is well aligned with the task of detecting archaeological structures from image data \citep{bundzel_semantic_2020,bonhage_modified_2021,davis_deep_2021,guyot_combined_2021,altaweel_automated_2022,banasiak_semantic_2022,fisher_multidisciplinary_2022}. MR-CNNs segment images via masked regions, aiding structure identification and validation. Furthermore, the emergence of transfer learning has amplified the use of CNNs in archaeology \citep{gallwey_bringing_2019,soroush_deep_2020,herrault_automated_2021}. By using a neural network model trained for a different but similar type of image data (\textit{e.g.} to detect bicycles in an image), and training it on the actual desired data (\textit{e.g.} detect motorcycles) but with a much smaller dataset, compared to the dataset needed without transfer learning, transfer learning can be helpful where only small datasets are available, reducing thus also the time and cost of data acquisition and preparation. In our dataset of articles, the use of transfer learning – counted at 41\% (n = 16) of all study cases reviewed on automatic structure detection - were generally based on models pre-trained with image datasets such as \textit{ImageNet} \citep{russakovsky_imagenet_2015} or \textit{COCO} \citep{lin_microsoft_2014}, though in one unique case authors also relied on specialised datasets \citep{silburt_lunar_2019}.

\par However, as Herrault et al. [\citeyear{herrault_automated_2021}] has highlighted, CNNs have limitations in their interpretability? and are also too sensitive for unbalanced classes (\textit{i.e.} the categories available for prediction), a common issue for archaeological data. A deeper exploration of models, as done by Monna et al. [\citeyear{monna_machine_2020}], exploring six distinct families of methods (ANNs, Bayesian classifiers, linear classifiers, ensemble learning, unsupervised learning and nearest neighbour classifiers), allows for a better understanding of the data, despite the increase in complexity and time. Following the training of the models, Monna et al. [\citeyear{monna_machine_2020}] aggregate predictions through a hard voting mechanism, itself ensemble learning, ultimately identifying random forest (an ensemble learning method) as the best model.

\par To evaluate the success of machine learning archaeological structure detection models, authors have access to a wide range of metrics. The F1-score is the most widely used with 20 accounted uses [\textit{e.g.} \citealp{character_archaeologic_2021, altaweel_automated_2022, banasiak_semantic_2022}], but prediction accuracy is also common, with 16 applications in the study cases on automatic structure detection [\textit{e.g.} \citealt{monna_machine_2020, davis_locating_2021}]. While not always directly discussed, precision, average precision, and recall can nevertheless be found in the results in most of the study cases reviewed here \citep{bonhage_modified_2021, berganzo-besga_hybrid_2021}. Intersect over union (IoU) is only present in five study cases \citep{bundzel_semantic_2020, bordon_automatic_2021,banasiak_semantic_2022,trotter_machine_2022,yang_auto-identification_2022}, despite revealing valuable information on a model accuracy and being particularly well-suited for image segmentation tasks. The newly developed “boundary IoU” \citep{cheng_boundary_2021} was specifically created for evaluating image segmentation models, and could become an important addition to future studies. New metrics have even been developed explicitly for archaeological image segmentation. Fiorucci et al. [\citeyear{fiorucci_deep_2022}] introduced centroid-based and pixel-based measures, as they found IoU not ideal for evaluating discrete archaeological objects.

\par The significant popularity of machine learning applications for automatic structure detection can be further found in its perceived suitability to answer the research question of whether a certain feature is an archaeological site or structure?. Machine learning methods seem to be particularly effective when applied to massive structures such as burial mounds \citep{guyot_detecting_2018, caspari_convolutional_2019, monna_machine_2020, berganzo-besga_hybrid_2021} or large-size archaeological sites \citep{menze_detection_2006, menze_mapping_2012, stott_searching_2019}).

\par To summarise, following the explosion of big data such as the use of archaeological remote sensing imagery (especially LiDAR), combined with the development of powerful deep learning models, new possibilities for the identification of settlements and archaeological structures have appeared. Leaning massively on transfer learning and pre-trained models, the aforementioned studies applying these newly developed models provide encouraging results to archaeologists. However, the diversity of applications, metrics, pre-processing, and the very nature of the neural network models all lead to a heterogeneity of methodologies available for this task.

\subsubsection{Digital heritage}
In the articles applying machine learning methods to digital heritage, we identified two different approaches. On the one hand, there were the study cases with a complete workflow, with a first step of semantic image segmentation preprocessing performed via machine learning before a second step of classification based on machine learning \citep{grilli_classification_2019, nogales_arqgan_2021}. These successive steps follow a gradual and systematic process, which meets the requirements for the \textit{Knowledge Discovery Process} [cf. \ref{4.2.1}, \citealt{fayyad_data_1996}]. On the other hand, there were works that focused either only on the semantic segmentation step \citep{felicetti_mose_2021, matrone_transfer_2021} or only on the classification step \citep{toler-franklin_multi-feature_2010, mesanza-moraza_machine_2021,prasomphan_toward_2022, pavan_kumar_image_2022, pepe_data_2022}.

\par The high number of ANN models for semantic image segmentation adapted to cultural heritage tasks confirms a tendency \citep{sultana_evolution_2020}, with ANNs generally obtaining better results compared to decision trees \citep{boston_comparing_2022}. Non-standardised datasets and the diversity in the preprocessing stages made for a heterogeneous assemblage of reviewed study cases, making them difficult to adequately compare. The lack of standard protocols \citep[part. 7]{fiorucci_machine_2020} has led to the continued use of older models [\textit{e.g.} \citealt{pepe_data_2022}]. Limiting the use of colour images or reducing the number of classes has been suggested to improve future models \citep{grilli_classification_2019}.

\par Two different approaches are currently used in machine learning techniques for digital heritage problems: one emphasises a more progressive and integrative workflow with, successive detailed pre-processing steps, while the other focuses on semantic segmentation. Although deep learning models, in particular ANNs, appear to be the most used models, results are still hardly trustworthy in most cases, due to the lack of comparative datasets and standardised metrics.

\subsubsection{Text analysis}
Text analysis is one of the earliest topics to benefit from machine learning applications \citep{boon_digital_2009}, as the potential benefits were quickly identified \citep{richards_text_2015}. The quantity of information and data collected in the archaeological field rose exponentially following the development of new technologies, systematic recording processes, and the development of rescue archaeology \citep[p. 229]{brandsen_information_2023}. This veritable deluge of records \citep{bevan_data_2015} has made analysing the tremendous number of archaeological texts difficult. It is therefore unsurprising that machine learning methods already developed for text extraction in other fields found adoption in archaeology here.

\par Two approaches have emerged from this new demand for parsing through large numbers of archaeological reports, along with a third unrelated one. Firstly, machine learning methods can quickly find similarity between two texts where a human may only find heterogeneity \citep{boon_digital_2009}. Secondly, machine learning can also be useful for highlighting the underrepresentation of certain archaeological findings in existing literature [\textit{e.g.} cremation: \citealt[p.6]{brandsen_burning_2021}]. Lastly, researchers have also used machine learning to extract and interpret graphemes or other signs from archaeological material. This last application focuses on extracting glyphs from images \citep{dhivya_tamizhi_2021} or finding matching pieces from document fragments \citep{abitbol_machine_2021}. For these image-based tasks, ANNs are the family of models used \citep{dhivya_tamizhi_2021,abitbol_machine_2021} likely due to their effectiveness in computer vision tasks (cf. \ref{4.2.1}). However, we consider the results to still be too limited as of now.

\par All study cases that used natural language processing, required for analysing content similarity in a text document, all relied on ANNs, except for the early application of Bonn et al. [\citeyear{boon_digital_2009}], which used instead an unsupervised memory-based-learning model (MBL). Large language models, having been introduced only recently \citep{vaswani_attention_2017}, did not feature in our dataset due to our cutoff point of 2022, though there have already been some applications more recently \citep{chang_survey_2024}. Despite their youth, the impact of large language models across modern society has been large \cite{tamkin_understanding_2021,eloundou_gpts_2023,clusmann_future_2023,tayan_considerations_2024} and the archaeological field will undoubtedly implement these new models for either research purposes or education \citep{agapiou_interacting_2023, cobb_large_2023}.

\par Despite its essential role in archaeological research, machine learning applications for text analysis are poorly represented in our corpus. Almost only relying on ANN models in our corpus, these applications face strong methodological and technical issues during the pre-processing of the textual data. However, the recent development of large language models, absent in our corpus, is likely to drastically change the automated analysis of archaeological textual data in the future.

\subsubsection{Taphonomic classification}
\label{4.2.6}
Despite the wide range of methods applied, many of the articles reviewed were found to contain important methodological issues, both from unclear and insufficient reporting of the methods applied, to important considerations of the data used that were seemingly eschewed.

\par Some of these concerns have already been highlighted [\citealp{mcpherron_machine_2022}; but see the reply by \citealt{abellan_high-accuracy_2022}, including its Supplementary Texts 1 and 2; \citealt{moclan_are_2023}; see also \citealt{courtenay_deep_2024}], specifically the unclear methodology for the process of "bootstrapping” described in one publication in our reviewed dataset [\citealt{dominguez-rodrigo_successful_2018}; but also present in our dataset in \citealt{dominguez-rodrigo_distinguishing_2018, aramendi_who_2019, moclan_classifying_2019, moclan_identifying_2020}; and \citealt{courtenay_combining_2019}]. The unclear methodology reported in Domínguez-Rodrigo [\citeyear{dominguez-rodrigo_successful_2018}] could be interpreted as stating that bootstrapping was applied to create a larger overall dataset which was then split for training and testing (as seems to also be explicitly stated in \citealt{dominguez-rodrigo_distinguishing_2018}; but less so in \citealt{aramendi_who_2019},  or \citealt{courtenay_combining_2019}]. This constitutes a methodological failure, as the models will share the same data for across both training and testing datasets, rendering any results unusable in the same way it would be to have an answer sheet when taking an exam. Although we do not wish to reiterate the “bootstrapping” discussion, it is still important to highlight how imprecise reporting of methodology and code unavailability can cast doubt of the validity of the methods applied.

\par When the data and code are made available, such as in Abellán et al. [\citeyear{abellan_high-accuracy_2022}], concerns are easier to both discard and confirm. For instance, the Ensemble.ipynb file provided in the Supplementary Data (henceforth SD) 2 contains an off-by-one error, which has the effect of excluding the last variable when training the models used. If this is indeed the code used for the main article (it is at least for the Supplementary Text 4), it would have considerable implications for the reported results and the accuracy of the reported methods. Moreover, the \texttt{R} file croc paper 2022 code.r contains additional problems for the “correct” bootstrapping section (module 4 in the file), and we could not run the code for this module without errors (with \texttt{R 4.3.2}, and \texttt{caret 7.0-1}).

\par On the other hand, the code suggests that the definition of "bootstrapping” used by the authors is to resample input data with replacement to obtain a larger synthetic input dataset and then performing k-folds when training the model (perhaps explaining the reported use of both bootstrapping and k-folds in \citealt{dominguez-rodrigo_successful_2018} and \citealt{dominguez-rodrigo_distinguishing_2018}). This is contrary to the ideal implementation of bootstrapping, where the resampling is done by each classifier in the ensemble learning model, thus creating multiple bootstrapped datasets (one per classifier), as well as providing an out-of-bag error to verify the accuracy of each classifier when aggregating the models \citep{breiman_bagging_1996}. The code also implies that their "proper bootstrapping” process for creating a larger dataset may also be applied to the testing data after splitting as well, but this would severely skew model accuracy metrics, since the same data could be correctly or incorrectly predicted multiple times.

\par Other issues may require even more technical knowledge to identify, such as those in Moclán et al. [\citeyear{moclan_classifying_2019}], who evaluated their models with both a bootstrapped and non-bootstrapped experimental sample, seeking to predict the agent (anthropic direct percussion, spotted hyena (\textit{C. crocuta}) carnivory and wolf (\textit{C. lupus}) carnivory) of bone fracture planes in bone fragments (Moclán et al. 2019). However, the dataset provided in the Electronic supplementary material 1 (ESM 1) also contains information on the bone fragment the fracture planes come from (\textit{e.g.} epiphysis presence, length, and number of planes), rather than only the information of the plane itself \citep[app. ESM 1]{moclan_classifying_2019}. In some cases a single bone fragment could have six fracture planes in the dataset, so when splitting the dataset, some input data may have come from the same bone as some testing data, and thus have this shared bone metadata, likely leading to overfitting as all bone fracture planes from the same bone fragment were caused by the same agent [see \citep{varoquaux_machine_2022}]. Though no code was made available, the authors do state that these metadata variables were indeed used \citep[p. 4668]{moclan_classifying_2019}. In addition, the number of entries in ESM 1 do not match the number of fragments reported for any of the three classes. Even when counting only unique bone fragments instead of planes using ESM 1, the numbers still do not match for any class. Furthermore, for the hyena class, it is also clear that some fragments did not have all their fracture planes added to the ESM 1 dataset, since there are an odd number of entries that state the planes per bone fragment is 2, thus implying there is at least one plane that was not included \citep{moclan_classifying_2019}.

\par The issues in Moclán et al. [\citeyear{moclan_classifying_2019}] are likely also present in another article [\citealt{moclan_identifying_2020}; and perhaps by \citealt{abellan_high-accuracy_2022}, at least for the \textit{tramp\_cm\_croc\_MDR\_EB.txt} dataset, which contains a "number of grooves" variable). Although the data and code used by Moclán et al. 2020 was not made public, the authors state that the work was a continuation of Moclán et al. [\citeyear{moclan_classifying_2019}], repeatedly citing it as their methodological starting point, and likely using its trained model, especially as the variables described were identical. Moclán et al. [\citeyear{moclan_identifying_2020}] focused on classifying bone fracture agency in archaeological – as opposed to experimental – material, They provide strong conclusions based on the results of a model with considerable unaddressed (and unacknowledged) problems. A similar issue could be raised for Abellán et al. [\citeyear{abellan_high-accuracy_2022}], though they are far more couched in their language. Nevertheless, Fig. 4 in Abellán et al. [\citeyear{abellan_high-accuracy_2022}] states that for some of the archaeological BSMs analysed, the models gave a classification probability of 99\% for being crocodile-made, but at the same time also a 99\% probability for being a cut mark. Oddly, the probabilities do not add up to 1 except for the last BSM shown (BOU-VP-11/12), the only one classified as a cut mark. Whether this was a transcribing error or a deeper issue is unclear, as detailed results are not available in the SD. The vast class imbalance of input data is another issue, despite the authors suggesting otherwise. Even classifying no crocodile BSMs correctly would provide an F1 score of 0.92 if only 11 out of 146 testing cut marks were misclassified, thus making the lack of confusion matrix or detailed results reporting concerning, even if no strong conclusion was put forth [see \citealt{courtenay_deep_2024}].

\par The use of machine learning methods to draw any conclusion from its own predictions of archaeological data must be very carefully done, the limitations of the methods must be clearly and transparently examined, and the strength of the conclusions must be scrupulously tempered. Anything else should not be considered good scientific practice.\\
However, despite the lengthy discussion above, not all study cases reviewed had methodological shortcomings. The methods used in other articles in this category such as Byeon et al. [\citeyear{byeon_automated_2019}] and Cifuentes-Alcobendas and Domínguez-Rodrigo [\citeyear{cifuentes-alcobendas_deep_2019}] are promising and well-grounded [with only a few caveats, such as the latter’s omission on whether BSMs from the same bone were split into both training and testing, which could have unduly increased validation accuracy; see also \citealt{courtenay_deep_2024}] and should definitely be the basis of further studies on the subject, including those seeking to independently verify and replicate the results of the numerous BSM agent classification studies, especially as recent attempts have not obtained the same optimistic results, despite employing robust methodology \citep{courtenay_deep_2024}.

\subsection{Recommendations and good practices}
From our comprehensive review, several lessons for future improvement have emerged, which we have categorised into either methodological or theoretical considerations.

\par A first significant methodological concern is the lack of standardised workflows and practices. As an example, for automatic structure detection tasks \citep{bellat_automated_2024}, it is impractical to compare studies with different inputs (\textit{e.g.} RGB, DEM, multispectral images) or different and unharmonised preprocessing. One step to increase some standardisation could be to promote transfer learning from pre-trained models based on publicly available datasets, which has proven effective in some reviewed study cases \citep{gallwey_bringing_2019,herrault_automated_2021}. It could be particularly efficient in artefact classification, cultural heritage reconstruction, or automatic structure detection, where numerous image collections already very similar in nature exist; \textit{e.g. BigEarthNet} ,\citep{sumbul_bigearthnet_2019} \textit{AID} \citep{xia_aid_2017}, \textit{Million AID} \citep{long_creating_2021}, and \textit{NWPU-RESISC45} \citep{cheng_remote_2017}. These collections demonstrated their utility in enhancing model accuracy for remote sensing tasks \citep{wang_empirical_2023,thapa_deep_2023}. For text extraction or grey literature comparison tasks, adopting a common workflow for data cleansing and preparation is crucial. In addition, reported metrics should be standardised to allow for better comparisons across studies. In the case of prediction problems, the metrics of precision and recall, as well as confusion matrices should always be available to the public. The field of isoscape analyses, where most publications follow a similar workflow and script based on Bataille et al. [\citeyear{bataille_bioavailable_2018,bataille_triple_2021}], provides a good example \citep{serna_implications_2020,janzen_spatial_2020,barberena_bioavailable_2021,holt_strontium_2021,bataille_triple_2021} fostering a community of practitioners rather than individual practices from isolated individuals. This approach could be a goal for different archaeological subfields aiming to develop machine learning applications and create a cohesive community. In their study, Batist and Roe [\citeyear[Table 5, Fig. 6]{batist_open_2024}] highlighted a similar phenomenon in an open archaeological dataset for specific communities (radiocarbon, database) that have very strong ties and an active and a solid sharing policy.

\par Another major concern is the difficulty of clearly delimiting the extent of machine learning. The definition used here was particularly strict, excluding algorithms that are more traditional statistical methods or mainly deal with dimensionality reduction, even if they were reported in many of the articles reviewed here as indeed machine learning, such as k-means, PAM/k-median \citep{mircea_sex_2015,altaweel_finding_2019,febriawan_detection_2020,cacciari_contribution_2021,demjan_laser-aided_2022,bouzid_towards_2022,fernee_rolling_2022,badawy_formation_2022} as well as PCA, LDA, and QDA \citep{monna_machine_2020,ma_aminoisoscapes_2021,abellan_high-accuracy_2022,anglisano_supervised_2022,badawy_formation_2022}. It is not easy to draw a line between what is machine learning and what is not, since it extends across the fields of both statistics and computer science, if not more \citep{bzdok_classical_2017,bzdok_statistics_2018}. Therefore, we do not wish to suggest our definition is conclusive, but we must also highlight the varied understanding of the concept across authors, which can make an analysis of machine learning applications in a specific field challenging.

\par Data availability and open access to data are often discussed as big challenges in archaeological research. In general, the entire workflow, from original data collection to publication, should be accessible. Following the FAIR principles \citep{wilkinson_fair_2016}, data can be stored in open access data articles (\textit{e.g. Journal of Open Archaeology Data}) or in institutional and international platforms (\textit{e.g. Zenodo, OSF, FigShare}). The code should also be accessible either on an online platform (\textit{e.g.GitHub, GitLab}) or as supplementary materials. Moreover, all the results, including those not discussed, should be available to the reader. Adopting a FAIR workflow and open-access data gives the opportunity to share methodologies and outcomes to a wider public, and potentially aid in the preservation of cultural heritage \citep{fisher_ethical_2021}. The \textit{Peer Community Journal Archaeology} (\textit{PCI Archaeology}) provides a good example of a fully accessible and FAIR process, requesting the open publication of all data, scripts, and code used for the published study, as well as a clear and thorough reporting of all the methods used, enough to be independently reproducible. A list of software and platforms adapted for increasing reproducibility in archaeological research, although slightly outdated, is given in supplementary tables of Strupler and Wilkinson [\citeyear{strupler_reproducibility_2017}].

\par A major theoretical issue in many studies is the absence of clear archaeological questions that require, or at least are well-suited to, machine learning methodologies to be answered. Authors develop proofs-of-concept without clearly stating the problems that necessitate the application of machine learning \citep{albertini_designing_2017, stott_searching_2019, gallwey_bringing_2019,ramazzotti_modeling_2020, ushizima_materials_2020, vos_model_2021, dhivya_tamizhi_2021, lyons_lidar_2022}. This issue was already noted in remote sensing applications in African archaeology: “\textit{Much of the recent literature employing new analytical methods for remote sensing is purely experimental and thus is interested solely in developing methods that can be more widely applied by future work}” \citep{davis_aerial_2020}. Clear, well-defined research questions are essential before applying machine learning methods, as simpler statistical solutions may suffice in many cases. The \textit{theory in, theory out}” concept developed by Radford and Joseph [\citeyear[Fig. 1]{radford_theory_2020}] meets all these requirements, with a theoretical statement prior to the model design and a reflection on the model’s efficiency and its limitations for future theoretical implementation after the model has been run. This gap between research questions (and also theory) and data interpretation is important not only for digital applications but more broadly to all analyses performed on archaeological data, as has been underlined by Perreault [\citeyear{perreault_quality_2019}].

\par Furthermore, the validity and (especially inter-rater) reliability of the datasets used are in many cases not questioned enough \citep{tennie_earliest_2023}, while it is an essential element for reproducibility. Another issue is noise, the outcomes may be valid and reliable if analysed together, but may have low precision. All these problems will become more urgent once a larger transition to machine learning-based methods has happened and no traditional benchmarks exist to compare the results obtained from the models to a broader archaeological context. Besides the fact that the traditional benchmarks are often poor in validity and reliability themselves, the opacity (cf. \textit{black box} in part \ref{4.1}) of many machine learning model families (especially artificial neural networks, which have recently surged in popularity) renders the whole problem even more difficult to solve. One key solution could reside in the development of explainable artificial intelligence (XAI), which intends to bring more transparency throughout the entire automation process \citep{barredo_arrieta_explainable_2020}.

\par
\begin{sidewaysfigure}[p]
	\centering
	\includegraphics[scale=2, trim = {4cm, 3cm, 0cm, 1cm}, clip]{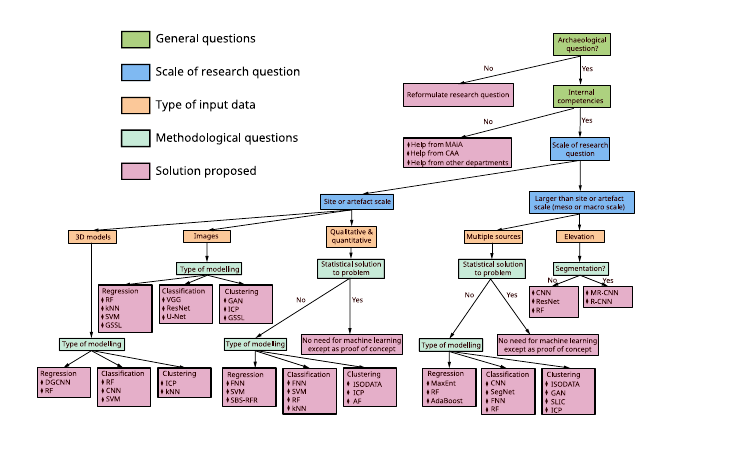}
\caption{Workflow adapted to machine learning solutions applied to archaeological problematics. Figure generated with Microsoft Word and Inkscape.}
	\label{fig11}
\end{sidewaysfigure}

 Interdisciplinarity and collaboration pose another challenge. Archaeologists should have the primary say in formulating research questions, even when working with computer scientists \citep[p. 15]{orton_mathematics_1980}. For example, the archaeology of early modern buildings integrates archaeologists and architects, yet the research questions ought to remain archaeological in nature for an archaeological study. Methodological questions can be developed by computer scientists, but the primary goals must stay aligned with archaeological objectives [\textbf{\cref{fig11}}].

\par Another important theoretical issue is the reflection throughout the automation process, which also corresponds to the concept of “\textit{theory in, theory out}”. As mentioned earlier, the use of the \textit{Knowledge Discovery Process} \citep{horr_algorithmen_2011, horr_machine_2014} represents an innovative and holistic approach to fully understand an archaeological artefact. It involves an initial clustering or unsupervised approach, followed by a supervised approach, with an optional semi-supervised step in between. This process encourages deeper reflection, a more subjective approach and more informed decision making. It is also a matter of taking the time to label the various data, as Klassen et al. state: “\textit{Given the nature of archaeological data, it is often difficult or expensive to get “labels” for things like artifact typologies and site chronologies}” \citep{klassen_semi-supervised_2018}. Therefore, standardised ontologies and semantic consistency are needed to assure the interoperability of the methods and results \citep{huggett_apparatus_2017, davis_defining_2020}.

\par Finally, while researchers can have doubts about the validity of results obtained through some applications of machine learning, as well as the validity of its underlying reasoning and the validity of the way these methods were applied [see \citealt{varoquaux_machine_2022}], and all these aspects need to be critically examined, they also have great potential. They can reveal unseen relationships within a dataset, moving beyond human interpretation to a more pragmatic system. As Ramazzotti [\citeyear[p. 174]{ramazzotti_modeling_2020}] states, machine learning in archaeology can recreate “a possible world of other associations of meaning devoid of sources and dispersed information, it exhibits the nuances and complex interrelations and, furthermore, it helps the interpreter codify other associations that were unforeseen (or hidden)”. Some might even consider highly mechanised and automated processes (machine learning being only one of many) as a new step into human technical gesture \citep[p. 74]{leroi-gourhan_geste_2022}. 

\clearpage
\newpage
\clearpage
\newpage

\subsection{Trends and perspectives}
Machine learning evolves rapidly, and the research presented in this article was based on articles published before 2023. New algorithms and applications that are not covered in our review have already emerged in the field of archaeology [\textit{e.g.} LLMs, cf. \citealt{agapiou_interacting_2023, cobb_large_2023, lapp_evaluating_2024}]. Any exhaustive literature review takes considerable time to accomplish, and it is difficult to remain up to date to a set of technologies so quickly evolving as is machine learning. Nonetheless, our review highlights several potential trends for the future of machine learning in archaeology.

\par Firstly, regarding the general development of machine learning in archaeology, the academic background of new students will evolve. We identified four programs that focused on computer science applications in archaeology. The “Computational Archaeology: GIS, Data Science and Complexity” master at UCL in London, the “Digital Archaeology” master at York University, the “Digital and Computational Archaeology” programme at Universität zu Köln in Cologne and the “Master in Digital Heritage and Landscape Archaeology” at the University of Cyprus in Nicosia. This number will undoubtedly grow, and machine learning will become as familiar to the next generation of archaeologists as GIS and remote sensing \citep{agapiou_remote_2015} or open source principles \citep{batist_open_2024} are today. Although not explored in this review, explainable artificial intelligence is likely to be trialled for many applications, as it also enables the archaeologist to see the hidden relationships between their data, processing workflow, and results. Archaeological development of such tools can allow more interpretability either during the process \citep{labba_iarch_2023} or \textit{post hoc} \citep{tenzer_debating_2024}.

\par Secondly, for image recognition tasks, particularly those related to cultural heritage and architecture, image segmentation will become an essential preprocessing step \citep{grilli_supervised_2018}. While already widely used, integrating segmentation at the beginning of a pipeline and linking it to a later classification algorithm is rare \citep{grilli_classification_2019, nogales_arqgan_2021}. This remark is also valid for any automatic structure detection tasks, but the large use of MR-CNNs suggests that many of the practitioners are already aware of the importance and usefulness of semantic image segmentation.

\par Regarding automatic structure detection, it seems that, without a doubt, economic opportunities will emerge for such applications. Private and public companies seek to manage costs and risks associated with new installations and construction, and automatic structure detection can help avoid unnecessary excavations. A private company in Ireland has already tested the \textit{ADAF} model \citep{coz_earthobservationadaf_2024} to reduce archaeological intervention time during track construction, by predicting unseen and unreferenced archaeological mounds. Archaeological predictive models might also be regarded as potential tools for construction companies or even public administration to regulate their cultural heritage conservation policy. However, it is crucial to use these tools responsibly, acknowledging that machine learning models are not infallible, nor can they replace manual fieldwork. These tools, if applied commercially, ought to be seen as an avenue to minimise or at the very least better predict the impact of archaeological material during construction, but never as a way to avoid archaeological intervention. In fact, it might be possible to observe or even prevent the destruction of archaeological material \citep{el-hajj_interferometric_2021}, especially with widely-verified and trusted methodologies. Both software developers and archaeologists must clearly define thresholds for feature classification or prediction and provide diverse datasets to ensure comprehensive predictions to avoid overlooking archaeological features.

\par The future development of new algorithms for classification of artefacts is another key area. Unlike automatic structure detection, where MR-CNNs dominate, there are no standard models in artefact classification. Although ANNs have taken the lead, as they have in other disciplines \citep{zheng_development_2021, osco_review_2021, thai_machine_2022}, a few dominant models, whether within the ANN family of methods or not, will likely emerge. \textit{ResNet} is already very popular and well-suited for image recognition problems \citep{gualandi_open_2021, anichini_automatic_2021, kowlessar_reconstructing_2021, berganzo-besga_hybrid_2021, jalandoni_use_2022}, despite its relative antiquity \citep{he_deep_2016}. Furthermore, the rising amount of data used will favour deep learning models compared to other types of models (\textit{e.g.} ensemble learning, Bayesian classifier), due to its higher number of parameters \citep{sarker_deep_2021}. 

\par In the case of APMs, ensemble learning methods like tuned RF and \textit{MaxEnt}, despite being several decades old, are well-suited to APM (\textit{MaxEnt} in particular), which does not require data points for absence of features \citep{yaworsky_advancing_2020, yaworsky_effects_2024, yaworsky_neanderthal_2024}. These methods will likely remain in use for some years due to the special situation of APMs. An alternative is to employ deep learning models coupled with interpretability tools such as \textit{LIME} \citep{ribeiro_why_2016} or \textit{SHAP} \citep{lundberg_unified_2017}, which were not applied in our corpus, and (to our knowledge) have yet to be applied in archaeology. One likely solution will be the development of studies combining \textit{"traditional"} statistical approaches with supervised or unsupervised machine learning approaches to hopefully obtain more readable and interpretable results \citep{li_gis_2024}.

\par Finally, we have not explored large language models (LLMs) in detail. The release of \textit{GPT-3.5} in 2022 opened many new possibilities \citep{brown_language_2020}. Our review did not cover these developments as we excluded any articles published after 2022. However, LLMs have already shown promising results in archaeology for teaching \citep{cobb_large_2023} and literature scraping, as well as classification \citep{agapiou_interacting_2023, bellat_fail_2024}. LLMs have even been applied to the automatic classification of artefacts with partially successful results \citep{lapp_evaluating_2024}. In the future, LLMs could also be used for code generation or translation between programming languages. While \textit{Unnatural-Code-LLaMA-34B} excels in generating new code, \textit{GPT-4} is the better option for code translation \citep{zheng_survey_2024}. LLMs could make coding more accessible to a wider audience of archaeologists and are likely to increase the number of publications on machine learning applications in archaeology. They are not without their concerns, however, as many academic institutions \citep{university_of_tubingen_guidelines_2023, oxford_university_guidelines_2024, massachusetts_institute_of_technology_initial_2024, wang_generative_2024} or even European deciders \citep{european_commission_living_2024} have already set up guidelines to control and regulate the use of such models.

\section{Conclusions}

Our extensive review of applications of machine learning in archaeological research, which includes both quantitative and qualitative observations, highlighted several key phenomena. Expanding on Bickler’s assertion that: “\textit{Machine learning arrives in archaeology}” \citep{bickler_machine_2021}, our findings suggest that machine learning has been widely present in archaeology since at least 2019. The adoption of machine learning techniques has been slower than in other fields, likely due to the inherent inertia of the discipline. To understand the diversity of machine learning approaches used in archaeology, we focused primarily on four parameters: archaeological subfields, family of algorithms applied, model evaluation goal, and scientific task. The initial wave of machine learning applications focused on artefact classification and bioarchaeological problems \citep{nguifo_plata_1997, toler-franklin_multi-feature_2010, ionescu_applying_2015, mircea_sex_2015}. Automatic structure detection has strongly developed after 2019, likely due to the proliferation of image recognition algorithms based on convolutional neural networks (CNNs) and segmentation models such as Multi-Scale Region-Based CNNs, especially since 2021. The heterogeneity of models used in archaeology contrasts with other fields such as medical imaging applications, where CNNs and generative adversarial networks (GANs) are predominantly used \citep{barragan-montero_artificial_2021}.

\par Our work also highlighted some methodological concerns likely arising from the limited background of many archaeologists in machine learning, as well as of some programmers in relation to archaeological problems. Davis and Douglass have already pointed out this disconnect in remote sensing applications in archaeology, noting a “\textit{disconnect between remote sensing applications and anthropological theory}” \citep{davis_aerial_2020}.  Another critical aspect is the reliability and variability of the benchmarks themselves, which are often poor, leading to additional problems in evaluating the models \citep{tennie_earliest_2023}. Furthermore, the suitability of the chosen machine learning methodology to answering the scientific question sought after is crucial: in many cases, statistical methods are sufficient to provide an answer, possibly rendering machine learning methods superfluous. To address the challenges of applying these methods to archaeological questions, we have designed a flowchart of suggested best practices [\textbf{\cref{fig11}}] to help archaeologists develop a coherent and effective approach. This small tool aims to bridge the gap between archaeological research and machine learning, promoting a more integrated and informed application of these technologies in the field.

\par Regarding the future of machine learning in archaeology, three key phenomena emerge. Firstly, archaeologists' ongoing training in digital methods will lead to a more comprehensive workflow and new applications in sub-fields of archaeology little represented in our current corpus (\textit{e.g.}, ethnoarchaeology, archaeobotany). Improving skills and knowledge will also reduce possible methodological issues, and provide a more direct interaction between the original research questions and the available avenues to solve them. Secondly, the constant development of new models will leave space for new proof-of-concept studies with more complex and wider datasets, but also allow for results that should improvement in results in some domains (cultural heritage, text analysis, automatic structure detection, and artefact classification). Thirdly, the recent boom in large language models and generative AI will change research in archaeology at different levels, as it does already for other fields of academia \citep{andersen_generative_2025}. Either by allowing the creation of helpful tools for data curation, teaching tools, or multivariate analysis, these models are easy to use and have already been adopted by many in their daily life (though ethical issues remain; see \citealt{hagendorff_mapping_2024}). One trend whose path remains uncertain is the possible formation of a reflexive community for AI and machine learning applications in archaeology (\textit{e.g.} such as the COST action \textit{MAiA}: \citealt{maia_managing_2025}; or CAA machine learning special interest group), leading to more standardised practices and protocols with ethical guidelines.

\par Despite the many concerns on applying machine learning in archaeology, these methods can be very powerful tools, able to massively reduce the time, cost, and (whether positive or not) labour to process, analyse, and predict even large-scale and complex data [\textit{e.g.} \citealt{kochkov_neural_2024}]. The existing reported successes of machine learning in other areas of science, and its already increasing number of applications in archaeology in recent years, all show it is cementing its place in the future of the field. We are likely at – if not already in a post-digital area \citep{gattiglia_managing_2025} –  a turning point in archaeology, as well as in other disciplines, where a new approach – machine learning –  has the possibility to permeate all levels of archaeological practice. Like any other statistical tool, machine learning itself is neither inherently good nor bad; its impact on the field of archaeology depends on how it is used \citep{huggett_apparatus_2017}. We should be mindful of the thinking derived from post-processual archaeology in the late 1970s \citep{hodder_reading_1986}, developed in response to processual archaeology \citep{phillips_method_1953}. The answer of this post-processual archaeology was to: “\textit{explain the past rather than describing it}” \citep{yoffee_archeologie_2010}. Following this perspective, machine learning applications in archaeology should never be limited to data processing or analysis but should be integrated into a broader archaeological reflection with precise questions, albeit knowingly and carefully, making sure its limitations are well known, well reported, and well addressed.

\textbf{\underline{Author contributions:}} M.B., J.D.O.F., C.T. and T.S. conceived the idea for the manuscript and initiated and designed the research. M.B. and J.D.O.F., collected data and selected the articles. M.B. and J.D.O.F., performed the statistical analysis and interpreted the results. M.B. and J.D.O.F. wrote the manuscript, with critical input from all co-authors. All authors reviewed the manuscript.

\par \textbf{\underline{Acknowledgment:}} We would like to thank J. Padarian for providing help on scraping strategy and Á. Tejero-Cantero for the generous help and patience afforded during the earlier stage of the project.

\par \textbf{\underline{Funding:}} This work has received funding from the Deutsche Forschungsgemeinschaft (DFG) Collaborative Research Center (CRC) 1070 “ResourceCultures”, grant agreement n°215859406 and from the European Research Council (ERC) under the European Union’s Horizon 2020 research and innovation program under grant agreement n°714658; (STONECULT project). We acknowledge support from the Open Access Publication Fund of the University of Tübingen.

\par \textbf{\underline{Conflict of interest statement:}} The authors declare no conflict of interest.

\label{annexe}
\par \textbf{\underline{Data availability statement:}} The data used in this study are openly available in an OSF repository at \url{https://doi.org/10.17605/OSF.IO/RUPGY}. Codes are on GitHub on “\href{https://github.com/ac-jorellanaf/Google-Scholar-Scraper}{Google-Scholar-Scraper}” for the manual screening and “ \href{https://github.com/mathias-bellat/ML_archaeology_review}{ML\_archaeology\_review}”. 

\par \textbf{\underline{Declaration of Generative AI and AI-assisted technologies in the writing process:}} During the preparation of this work the authors used \textit{ChatGPT 3.5} \citep{brown_language_2020} in order to generate part of the R code used for the automatic screening protocol searches and figure plotting. After using this tool, the authors reviewed and edited the content as needed and took full responsibility for the content of the publication.

\section{Annexes}
\begin{center}
\footnotesize
\setlength{\tabcolsep}{3pt} % ajuste l'espacement entre les colonnes

\begin{longtable}{%
    m{9cm}|
    m{2cm}|
    m{1.5cm}|
    m{2cm}|
}
\caption{List of algorithms present in the study cases reviewed, along with their abbreviations, number of uses in our corpus, and number of times they performed best in comparative evaluations.} \label{annexe1} \\
\toprule
Description & Abbreviation & Uses & Best performance \\
\midrule
\endfirsthead

\caption[]{(continued)}\\
\toprule
Description & Abbreviation & Uses & Best performance \\
\midrule
\endhead

\bottomrule
\endfoot

% --------------------------
% ARTIFICIAL NEURAL NETWORKS
% --------------------------
\multicolumn{4}{c}{\bfseries Artificial Neural Network} \\
\midrule

Feedforward Neural Network & FNN & 23 & 4\\
Convolutional Neural Network & CNN & 14 & 1\\
Residual Neural Network & ResNet & 12 & 2\\
Mask Region-based Convolutional Neural Network & MR-CNN & 9 & 1\\
Faster Region-based Convolutional Neural Network & FR-CNN & 8 & 0\\
Visual Geometry Group & VGG & 8 & 2\\
U-Net & U-Net & 7 & 4\\
Inception Network & INC & 4 & 1\\
AlexNet & AlexNet & 3 & 0\\
RetinaNet & RN & 3 & 0\\
YOLO & YOLO & 3 & 0\\
DeepLabv3+ & DL3 & 2 & 0\\
Semantic Segmentation Model & SegNet & 2 & 0\\
Adaptive deep learning for fine-grained image retrieval & ADLFIG & 1 & 0\\
Bidirectional Encoder Representations from Transformers (DNLM) & BERT & 1 & 0\\
BiGRU-Dual Attention & BiGRU & 1 & 0\\
BiLSTM (RNN/Recurrent Neural Network) & BiLSTM & 1 & 0\\
Dynamic Graph Convolutional Neural Network & DGCNN & 1 & 0\\
DenseNet201 & DN201 & 1 & 0\\
Generative Adversarial Network & GAN & 1 & 0\\
Jason 2 & JAS2 & 1 & 0\\
Neural Support Vector Machine & NSVM & 1 & 0\\
Region-based Convolutional Neural Network & R-CNN & 1 & 0\\
SimpleNet & SimpleNet & 1 & 0\\
Single Shot MultiBox Detector & SSD & 1 & 0\\

% --------------------------
% BAYESIAN CLASSIFIER
% --------------------------
\midrule
\multicolumn{4}{c}{\bfseries Bayesian Classifier} \\
\midrule

Naïve Bayes & NB & 11 & 0\\
Maximum Entropy & MaxEnt & 2 & 0\\

% --------------------------
% DECISION TREES AND RULE INDUCTION
% --------------------------
\midrule
\multicolumn{4}{c}{\bfseries Decision Trees and Rule Induction} \\
\midrule

C5.0 & C5.0 & 7 & 2\\
C4.5 & C4.5 & 4 & 0\\
Decision Tree / Classification Tree & DT & 4 & 0\\
Conditional Inference Trees & CTREE & 2 & 0\\
Iterative Dichotomiser 3 & ID3 & 2 & 0\\
Classification And Regression Tree & CART & 1 & 0\\
Fast and Frugal Tree & FFT & 1 & 0\\
Learning with Galois Lattice & LEGAL & 1 & 0\\
Representative Trees & REPTree & 1 & 0\\
Random Tree & RT & 1 & 0\\

% --------------------------
% ENSEMBLE LEARNING
% --------------------------
\midrule
\multicolumn{4}{c}{\bfseries Ensemble Learning} \\
\midrule

Random Forest & RF & 54 & 20\\
Adaptive Boost & AdaBoost & 2 & 0\\
Stochastic Gradient Boosting & SGB & 2 & 1\\
eXtreme Gradient Boosting & XGB & 2 & 1\\
Bootstrap Aggregating & BAgg & 1 & 0\\
Discrete SuperLearner & dSL & 1 & 0\\
Fast Random Forest & FRF & 1 & 0\\
Gradient Boosting Regression Tree & GboostRT & 1 & 0\\
LogitBoost & LB & 1 & 0\\
Quantile Random Forest & qRF & 1 & 0\\
Sequential Backward Selection–Random Forest Regression & SBS-RFR & 1 & 1\\
SMOTE Boost & SMOTEBoost & 1 & 0\\
SMOTE + Edited Nearest Neighbour Rule & SMOTEENN & 1 & 0\\
Super Learner & SP & 1 & 1\\
Viola-Jones Cascade Classifier & VL-CC & 1 & 0\\
Genetic Algorithm & GA & 1 & 0\\

% --------------------------
% LINEAR CLASSIFIER
% --------------------------
\midrule
\multicolumn{4}{c}{\bfseries Linear Classifier} \\
\midrule

Support Vector Machine & SVM & 26 & 2\\
Structured SVM & SSVM & 1 & 0\\

% --------------------------
% NEAREST NEIGHBOUR CLASSIFIER
% --------------------------
\midrule
\multicolumn{4}{c}{\bfseries Nearest Neighbour Classifier} \\
\midrule

k-nearest neighbors & kNN & 19 & 1\\
Weighted k-nearest neighbors & kkNN & 3 & 0\\

% --------------------------
% POLYNOMIAL CLASSIFIER
% --------------------------
\midrule
\multicolumn{4}{c}{\bfseries Polynomial Classifier} \\
\midrule

Support Vector Machine with Radial Basis Function Kernel & SVMr & 7 & 1\\

% --------------------------
% UNSUPERVISED LEARNING AND CLUSTERING
% --------------------------
\midrule
\multicolumn{4}{c}{\bfseries Unsupervised Learning and Clustering} \\
\midrule

Affinity Propagation & AF & 1 & 0\\
Hierarchical Cluster-Based Peak Alignment & CluPA & 1 & 0\\
Databionic Swarm & DBS & 1 & 0\\
Expectation-Maximisation Clustering & EMC & 1 & 0\\
Graph-based Semi-Supervised Learning & GSSL & 1 & 1\\
Iterative Closest Point & ICP & 1 & 0\\
Iterative Self-Organizing Data Analysis & ISODATA & 1 & 0\\
Nearest Centroid & NC & 1 & 0\\
Simple Linear Iterative Clustering & SLIC & 1 & 0\\
Self-Organizing Map & SOM & 1 & 0\\
Tilburg Memory-Based Learning & TiMBL & 1 & 0\\
Time series clustering & TSC & 1 & 0\\
k-Mean Clustering & k-MC & 7 & 0\\

\end{longtable}
\end{center}

\subsection{Glossary}
\begin{itemize}
    \item ALS = Airborne laser scanning
    \item APM = Archaeological predictive models
    \item G.I.S. = Geographic information system
    \item GPR = Ground penetrating radar
    \item ICP = Iterative closest point
    \item LDA = Linear discriminant analysis
    \item LiDAR = Light detection and ranging
    \item LogR = Logistic regression
    \item MCC = Matthews correlation coefficient
    \item MDA = Mixture discriminant analysis
    \item ML = Machine learning
    \item NLP = Natural language processing
    \item PCA = Principal component analysis
    \item QDA = Quadratic discriminant analysis
    \item ROC = Receiver operating characteristic
    \item UAV = Unmanned aerial vehicle
    \item XRF = X-ray fluorescence
\end{itemize}

\newpage

\bibliographystyle{unsrtnat} %unsrtnat

\end{document}